\PassOptionsToPackage{unicode}{hyperref}
\PassOptionsToPackage{hyphens}{url}
\PassOptionsToPackage{dvipsnames,svgnames,x11names}{xcolor}
\documentclass[
  11pt,
]{article}
\usepackage{amsmath,amssymb}
\usepackage{iftex}
\ifPDFTeX
  \usepackage[T1]{fontenc}
  \usepackage[utf8]{inputenc}
  \usepackage{textcomp} 
\else 
  \usepackage{unicode-math} 
  \defaultfontfeatures{Scale=MatchLowercase}
  \defaultfontfeatures[\rmfamily]{Ligatures=TeX,Scale=1}
\fi
\usepackage{lmodern}
\ifPDFTeX\else
\fi
\IfFileExists{upquote.sty}{\usepackage{upquote}}{}
\IfFileExists{microtype.sty}{
  \usepackage[]{microtype}
  \UseMicrotypeSet[protrusion]{basicmath} 
}{}
\usepackage{xcolor}
\usepackage[margin=1in]{geometry}
\usepackage{graphicx}
\makeatletter
\def\maxwidth{\ifdim\Gin@nat@width>\linewidth\linewidth\else\Gin@nat@width\fi}
\def\maxheight{\ifdim\Gin@nat@height>\textheight\textheight\else\Gin@nat@height\fi}
\makeatother
\setkeys{Gin}{width=\maxwidth,height=\maxheight,keepaspectratio}
\makeatletter
\def\fps@figure{htbp}
\makeatother
\newlength{\cslhangindent}
\newlength{\csllabelwidth}
\newlength{\cslentryspacingunit} 
\newenvironment{CSLReferences}[2] 
 {
  \setlength{\parindent}{0pt}
  \ifodd #1
  \let\oldpar\par
  \def\par{\hangindent=\cslhangindent\oldpar}
  \fi
  \setlength{\parskip}{#2\cslentryspacingunit}
 }%
 {}
\usepackage{calc}

\usepackage{accents}
\usepackage{amsmath}
\usepackage{bm}
\usepackage{booktabs}
\usepackage{braket}
\usepackage{caption}
\usepackage{dcolumn}
\usepackage{epigraph}
\usepackage{extarrows}
\usepackage{float}
\usepackage[bottom,flushmargin,hang,multiple]{footmisc}
\usepackage{graphicx}
\usepackage{hyperref}
\usepackage[export]{adjustbox}
\usepackage{libertine}
\usepackage[libertine]{newtxmath}
\usepackage[mathcal]{euscript}
\usepackage{mathrsfs}
\usepackage{multirow}
\usepackage{quoting}
\usepackage{scrextend}
\usepackage{setspace}
\usepackage{soul}
\usepackage{subcaption}
\usepackage{tabularx}
\usepackage{titlesec}
\usepackage[utf8]{inputenc}
\ifLuaTeX
  \usepackage{selnolig}  
\fi
\IfFileExists{bookmark.sty}{\usepackage{bookmark}}{\usepackage{hyperref}}
\IfFileExists{xurl.sty}{\usepackage{xurl}}{} 
\hypersetup{
  pdftitle={AI Knowledge and Reasoning: Emulating Expert Creativity in Scientific Research},
  colorlinks=true,
  linkcolor={Maroon},
  filecolor={Maroon},
  citecolor={Blue},
  urlcolor={blue},
  pdfcreator={LaTeX via pandoc}}

\title{AI Knowledge and Reasoning: Emulating Expert Creativity in
Scientific Research}
\author{}
\date{\vspace{-2.5em}}

\begin{document}
\maketitle

\quotingsetup{font={itshape}, leftmargin=2em, rightmargin=2em, vskip=1ex}
\begin{center}
\author
{Anirban Mukherjee,$^{1\ast}$ Hannah H. Chang$^{2}$\\
\medskip
\normalsize{$^{1}$Samuel Curtis Johnson Graduate School of Management, Cornell University,}\\
\normalsize{Sage Hall, Ithaca, NY 14850, USA}\\
\normalsize{$^{2}$Lee Kong Chian School of Business, Singapore Management University,}\\
\normalsize{50 Stamford Road, Singapore, 178899}\\
\smallskip
\normalsize{$^\ast$To whom correspondence should be addressed; E-mail: am253@cornell.edu.}\\
}
\end{center}
\medskip
\singlespacing

\begin{center} 
\noindent \textbf{Abstract}
\end{center}

\noindent We investigate whether modern AI can emulate expert creativity
in complex scientific endeavors. We introduce novel methodology that
utilizes original research articles published after the AI's training
cutoff, ensuring no prior exposure, mitigating concerns of rote
memorization and prior training. The AI are tasked with redacting
findings, predicting outcomes from redacted research, and assessing
prediction accuracy against reported results. Analysis on 589 published
studies in four leading psychology journals over a 28-month period,
showcase the AI's proficiency in understanding specialized research,
deductive reasoning, and evaluating evidentiary alignment---cognitive
hallmarks of human subject matter expertise and creativity. These
findings suggest the potential of general-purpose AI to transform
academia, with roles requiring knowledge-based creativity become
increasingly susceptible to technological substitution.

\begin{center}\rule{0.5\linewidth}{0.5pt}\end{center}


\newpage
\doublespacing

\hypertarget{introduction}{%
\section{Introduction}\label{introduction}}

Advances in artificial intelligence (AI) raise pressing questions about
technological displacement
(\protect\hyperlink{ref-stone2022artificial}{Stone et al. 2022}). A
long-standing view posits education shields specialized labor
(\protect\hyperlink{ref-frey2019technology}{Frey 2019}, see Figure 4 in
\protect\hyperlink{ref-royalty1998job}{Royalty 1998}). Specifically,
automation risks are concentrated among lower-skilled occupations
focused on routine manual tasks. For example, Acemoglu and Restrepo
(\protect\hyperlink{ref-acemoglu2022tasks}{2022}) found that ``between
50\% and 70\% of recent changes in the U.S. wage structure are accounted
for by routine task-specialized worker groups in rapidly automating
industries.'' In contrast, advanced training forms `moats' around
expertise-intensive fields like law and academia, protected by their
inherent complexity
(\protect\hyperlink{ref-brynjolfsson2014second}{Brynjolfsson and McAfee
2014}, \protect\hyperlink{ref-cowen2013average}{Cowen 2013}).

Central to this perspective is the role of knowledge-based
creativity---novel, useful, and therefore creative
(\protect\hyperlink{ref-amabile2011componential}{Amabile 2011})
problem-solving in fields requiring complex thought and specialized
training. A preeminence of creativity limits automation risk if the
ability of machines to engage in novel thought is challenged by their
characterization as `stochastic parrots'
(\protect\hyperlink{ref-bender2021dangers}{Bender et al. 2021}), merely
regurgitating information found in their training data. Indeed, if AI's
responses are solely based on likelihoods inferred from prior exposure
without any genuine understanding or consciousness, then novelty in
thought is, by construction, unfeasible---the model's outputs must be
simply those that best match the prior conditional likelihoods of
appearance, and therefore words, sentences, and documents that exist and
were used to train the model; the output must, by definition, be banal.

Recent evidence, however, suggests a singularity---a point at which AI
systems advance beyond human reasoning capabilities---thereby
challenging this assumption
(\protect\hyperlink{ref-kurzweil2005singularity}{Kurzweil 2005},
\protect\hyperlink{ref-lyytinen2003disruptive}{Lyytinen and Rose 2003}).
Contemporary AI is displaying unexpected aptitude in complex cognition
(\protect\hyperlink{ref-bommasani2021opportunities}{Bommasani et al.
2021}, \protect\hyperlink{ref-hendrycks2021measuring}{Hendrycks et al.
2021}) and in processing complex knowledge gleaned from large-scale
datasets, such as scientific articles and textbooks, suggesting
boundaries may begin to blur. For instance, a recent workforce survey
(\protect\hyperlink{ref-Gutierrez2023}{Gutierrez 2023}) revealed that
while 72\% of workers report increased productivity due to AI, 42\%
expressed concerns about the technology's impact on their jobs. Notably,
44\% of individual contributors reported being `very or somewhat
concerned,' compared to 38\% of managers or those in higher positions.
The perception of AI as a job threat also varied by salary: More workers
earning under \$50,000 a year were concerned (47\%) about the
technology's impact compared to those earning between \$50,000 and
\$99,000 (39\%) or \$100,000 or more annually (36\%)
(\protect\hyperlink{ref-Caminiti_2023}{Caminiti 2023}). Such
observations have led technology prognosticators to argue that
disparities in physical progress---with chatbots outpacing
robotics---may enable AI to gain a stronger foothold in high-income
knowledge professions, where motor skills are nonessential, before
affecting occupations like carpentry or masonry, where manual dexterity
is a key determinant (\protect\hyperlink{ref-billard2019trends}{Billard
and Kragic 2019}).

Pivotal to AI's capabilities is conceptual knowledge---the understanding
of ideas, principles, and categories that help us make sense of the
world (\protect\hyperlink{ref-mcrae201314}{McRae and Jones 2013},
\protect\hyperlink{ref-rips201211}{Rips et al. 2012}). Conceptual
knowledge encompasses the mental representations and structures we use
to organize and interpret information, enabling the recognition of
relationships among objects, events, and abstract ideas. It enables
individuals to categorize and classify experiences, apply general
principles to specific instances, and comprehend abstract concepts that
do not rely solely on sensory experience or direct observation. It
explains the `why' and `how' of things as they are.

Crucially, existing studies detailing AI's conceptual knowledge have
primarily examined more basic, everyday concepts, as is typical in
traditional human laboratory experiments. For instance, studies have
assessed inductive reasoning about commonplace categories like animals
or objects (\protect\hyperlink{ref-hayes2010inductive}{Hayes et al.
2010}, \protect\hyperlink{ref-lampinen2022can}{Lampinen et al. 2022}).
Prior studies of property induction in AI relied on simpler constructs
detached from specialized contexts. Han et al.
(\protect\hyperlink{ref-han2022human}{2022}) employed basic animal
inferences, such as ``people who are told that cats have some property
are more inclined to infer that similar animals like lions share that
property.'' Misra et al.
(\protect\hyperlink{ref-misra2022property}{2022}) leveraged
straightforward canonical properties such as ``a cat has whiskers,'' and
Binz and Schulz (\protect\hyperlink{ref-binz2023using}{2023}) utilized
textbook reasoning scenarios about ``Linda'' and ``Blickets.''

These tightly-controlled designs facilitate comparisons to human
cognition. However, human subject matter experts tackle intricate
theoretical knowledge and ambiguous, interconnected concepts in
practice; cognition in real-world problems relies heavily on integrating
intricate representations
(\protect\hyperlink{ref-hampton2006concepts}{Hampton 2006}). Prior
studies have overlooked AI's aptitude for understanding intricate
concepts and interdependencies that underlie domain expertise. Consider
the superficially synonymous terms `integration' in mathematics and
`mixing' in chemistry---despite similarities in the vernacular, grasping
their precise meanings within each field remains critical for experts.
Thus, while a narrow focus on basic concepts enables isolated
appraisals, an evidentiary gap persists concerning AI's capacity for
complex, context-dependent, and dynamic reasoning.

In this paper, we examine AI efficacy in perhaps the most complex of
human undertakings: the development of new scientific knowledge through
the scientific research process. Here, two factors---novelty and
conceptual complexity---play a pivotal role in the integration of
theoretical concepts from disparate fields into novel frames,
hypotheses, and findings
(\protect\hyperlink{ref-klahr2000exploring}{Klahr 2000}). We argue that
if AI exhibits fluidity in contextualized creative reasoning leading to
specialized knowledge, it would indicate progress beyond mere mechanical
statistical relationships towards adaptable cognition. Such progress may
threaten many expertise-intensive professions.

Fundamental to the measurement of creativity is placing the AI in novel
contexts and situations distinct from the examples it has seen during
its training and from its current understanding of complex conceptual
knowledge. Prior examinations of AI capabilities have been marred by
accusations of rote memorization stemming from the use of standardized
assessments and experimental setups in which information presented to
the AI for analysis is reworded or restructured but otherwise not new or
unique (\protect\hyperlink{ref-ullman2023large}{Ullman 2023}). What is
needed, therefore, are contexts and situations in which we can be
certain that the concepts build upon prior concepts and potentially the
AI but extend beyond to include new knowledge that the AI was not
exposed to, and whose development requires the creative use of prior
knowledge.

\hypertarget{aims-and-contributions}{%
\subsubsection{Aims and Contributions}\label{aims-and-contributions}}

Addressing this research gap, our paper introduces a novel methodology
specifically designed to assess AI's efficacy as a creative aide within
the scientific process, with a particular focus on contexts involving
new knowledge. We seek to address and mitigate issues of rote
memorization and to closely examine the AI's proficiency in conceptual
combination---the merging of two or more distinctive concepts to form a
new, often more complex, concept---challenges that significantly impact
the validity of existing research methods and findings.

To avoid rote memorization, we leverage the AI's training data cutoff
date; for instance, GPT-4's dataset was last updated in September 2021.
By focusing on research published after this date, we ensure that GPT-4
encounters material it has not been previously exposed to. Given that
GPT-4 was trained on a broad dataset, any demonstrated capabilities
would suggest a general aptitude for reasoning and learning, beyond mere
specialization. This methodology allows for an equitable analysis of the
AI's cognitive abilities when faced with new conceptual knowledge,
providing a grounded assessment of its creative capabilities. To our
knowledge, this experimental design is unique to our study.

Our approach contrasts with traditional approaches that have
predominantly measured AI performance through standardized tests, such
as the SAT, GRE, and domain-specific assessments like the AP Art History
exam (e.g., \protect\hyperlink{ref-achiam2023gpt}{Achiam et al. 2023},
\protect\hyperlink{ref-chang2023survey}{Chang et al. 2023}). However,
these methods often fall short in capturing the AI's ability to adapt to
real-world challenges, which require a dynamic and fluid integration of
concepts. For example, an AI preparing for medical licensing exams might
successfully memorize past tests and their solutions to pass. Yet, it
remains uncertain whether such narrowly focused preparation equips the
AI to effectively navigate the complexities of an ambiguous clinical
scenario. Similarly, while an AI might perform well on a bar exam after
training on study guides, its ability to apply legal concepts creatively
in unscripted scenarios presents a more significant challenge to its
core competencies. Our methodology is deliberately crafted to circumvent
these limitations by demanding that the AI engage with entirely novel
scenarios, encompassing both the content and the structure of the tests.

Other studies offer insights into AI's potential impact on various
professions by surveying humans and AI. Such studies either directly
solicit opinions on the extent to which AI can perform certain roles in
professions (e.g., \protect\hyperlink{ref-van2023ai}{Van Noorden and
Perkel 2023}) or aggregate opinions on the perceived efficacy of AI on
specific tasks to evaluate its capabilities in different occupations
(\protect\hyperlink{ref-eloundou2023gpts}{Eloundou et al. 2023}). While
these belief-based assessments contribute valuable perspectives, they
inherently reflect the limitations associated with survey data.
Specifically, such data capture expectations of AI's capabilities rather
than empirical evidence of its actual performance---a key distinction
given the recency of modern AI and the paucity of data on its actual
performance in such roles. Thus, reported beliefs are unlikely to be
data-driven and founded on actual metrics, raising the question of how a
human might assess what an AI is capable of if limited empirical
evidence exists of the AI's capabilities.

These contributions relate to the literature in the following ways.
First and foremost, they discuss the abilities of artificial agents to
engage in complex and creative thinking. By focusing on analytical
creativity, our work aligns with the call for research exploring how AI
navigates novel issues creatively---logically problem-solving in a
complex domain (\protect\hyperlink{ref-ding2020logical}{Ding 2020}).
This approach not only challenges the traditional boundaries of AI's
capabilities but also provides a practical framework for evaluating AI's
role in augmenting human creativity, particularly in fields that demand
high levels of innovation and novel thought
(\protect\hyperlink{ref-fugener2022cognitive}{Fügener et al. 2022}).
More broadly, they relate to a long standing tradition in information
systems and economics of measuring the productivity of information
technology (\protect\hyperlink{ref-acemoglu2011skills}{Acemoglu and
Autor 2011}, \protect\hyperlink{ref-brynjolfsson2022turing}{Brynjolfsson
2022}, \protect\hyperlink{ref-mithas2016information}{Mithas and Rust
2016}, \protect\hyperlink{ref-tambe2012productivity}{Tambe and Hitt
2012}), and its workforce consequences
(\protect\hyperlink{ref-genz2021workers}{Genz et al. 2021},
\protect\hyperlink{ref-peng2020does}{Peng and Zhang 2020}).

Second, we devise a novel process to test AI creativity. We simulate
critical stages of the scientific inquiry process to scrutinize the
model's potential to mirror expert creativity. We task an AI instance
with redacting findings from article abstracts while maintaining
contextual integrity---a task requiring linguistic precision and
conceptual understanding. A second instance assesses the completeness of
the redactions, gauging counterfactual reasoning skills. A third
instance predicts study outcomes based solely on the redacted research,
testing analytical acumen and grasp of underlying theories. Fourth and
fifth AI instances then compare the predictions to actual results to
determine proficiency in recognizing and evaluating the alignment of
findings. Jointly, these instances provide evidence on the extent to
which the AI is able to undertake creative tasks like a human, and
therefore the extent to which the AI can replace current human aides and
provide assistance in future scientific ventures.

Third, we present `in-silico' evidence of AI's creativity in the
real-world by analyzing 589 original research articles published between
October 2021 and January 2024 across four leading psychology journals:
\emph{Cognitive Psychology}, \emph{Journal of Experimental Psychology:
General}, \emph{Journal of Personality and Social Psychology}, and
\emph{Psychological Science}. What distinguishes psychology as an ideal
test-bed for our investigation is its focus on human subjects and the
AI's exposure to extensive data detailing human thoughts, behaviors, and
interactions. Consequently, it can harness both its explicit learnt and
implicit learnt knowledge. That is, when AI systems are trained on
textbooks, research articles, and other scholarly materials, and
evaluated using standardized tests like the SAT, they undergo explicit
learning. When they observe humans and formalize the observed patterns
into theory, they learn implicitly. The latter is germane to our
theorizing and tests as it provides a pathway for the development of new
knowledge, as would be essential for true creativity.

\hypertarget{illustration}{%
\paragraph{Illustration}\label{illustration}}

Consider the following excerpt from our data: ``Using a longitudinal,
epidemiological, clinical-pathologic cohort study of older adults in the
United States (N = 348), the present research investigated associations
between well-being and cognitive resilience. Consistent with
preregistered hypotheses, results showed that higher eudaimonic
well-being (measured via the Ryff Psychological Well-Being Scale) and
higher hedonic well-being (measured via the Satisfaction with Life
Scale) were associated with better-than-expected cognitive functioning
relative to one's neuropathological burden (i.e., beta-amyloid,
neurofibrillary tangles, Lewy bodies, vascular pathologies, hippocampal
sclerosis, and TDP-43).''

Consider its redacted variant where the empirical findings have been
obfuscated: ``Using a longitudinal, epidemiological, clinical-pathologic
cohort study of older adults in the United States (N = 348), the current
research examined possible associations between well-being and cognitive
resilience. The study considered both eudaimonic well-being (measured
via the Ryff Psychological Well-Being Scale) and hedonic well-being
(measured via the Satisfaction with Life Scale) in its investigation of
cognitive functioning in relation to one's neuropathological burden
(namely, beta-amyloid, neurofibrillary tangles, Lewy bodies, vascular
pathologies, hippocampal sclerosis, and TDP-43).''

To a trained psychologist, the research goal is apparent from the
redacted variant---exploring the relationship between eudaimonic and
hedonic well-being and cognitive functioning. However, the redacted
variant presents complex constructs and theories without specifying the
nature of the hypothesized relationships. To a reader unfamiliar with
these nuanced categories, other speculative connections may seem equally
plausible. For instance, one could infer the research examines the
relationship just as much between eudaimonic well-being and
neuropathological burden, or between cognitive functioning and
neuropathological burden, as between overall well-being and cognitive
resilience. Only a detailed understanding of these categories and their
intricate interrelationships would allow an expert to accurately
anticipate the study's hypothesized relationships and research design;
such expertise arises from a confluence of specialized knowledge and
sophisticated reasoning---without knowledge, one might engage in faulty
reasoning, and without reasoning, one cannot effectively apply
knowledge.

Note the contrast between the complexity of these constructs versus
everyday categories used in existing cognitive research. While such
studies often examine basic reasoning about animals, objects, and their
straightforward observable properties, constructs like well-being,
cognitive resilience, and neuropathological burden entail extensive
interrelated abstractions. Mastering these requires decades of
specialized education and employment for human experts. Can a publicly
available AI trained on a large-scale but general dataset emulate
similar capabilities?

Moreover, such concepts are embedded within a web of intricate
relationships requiring sophisticated theories to decipher. This
distinction highlights another key challenge: Can AI move beyond merely
recognizing isolated facts to actively navigating the nuanced fabric of
concepts defining academic discourse? Successfully parsing intricate
conceptual connections woven through scholarly writing would signal
meaningful progress in AI capability, as it would demonstrate aptitude
in engaging with the specialized knowledge that characterizes scientific
expertise. Assessing whether an AI can exhibit such in-depth
comprehension is a core goal of our proposed methodology.

\hypertarget{overview-of-results}{%
\paragraph{Overview of Results}\label{overview-of-results}}

Our results suggest that AI exhibits the capability to: (1) maintain the
integrity of the research context without revealing empirical findings,
a demanding cognitive task that would likely challenge even human
experts; (2) predict empirical outcomes of studies based on redacted
research; and (3) discern the theoretical implications of those
findings, mimicking an understanding of research context, question, and
design.

These findings apply to completely new research of which the AI is
unaware and has not been trained on. This focus on novelty is essential
in our study, as useful thought that is not novel cannot, by definition,
be creative. Similarly, novelty for the sake of novelty cannot lead to
creative problem-solving and its implications on technology adoption and
employment. For instance, were the AI aware of and able to reason a
proof of the Central Limit Theorem, by way of example, then such
reasoning would be evidence of its ability to emulate cognitive
knowledge but not its ability to demonstrate creativity, as the subject
matter is common and prevalent and therefore likely a part of its
training. In contrast, it is the novelty of the knowledge that the AI is
tested upon that constitutes evidence of its ability to deal with
novelty, whereby the AI is able to reason about new-to-it information.

Crucially, our findings contradict assumptions about general AIs lacking
aptitude for advanced reasoning. For instance, Goyal and Bengio
(\protect\hyperlink{ref-goyal2022inductive}{2022}) argue that machine
learning systems achieve narrow proficiency, unlike flexible human
cognition for generalizable understanding: ``Our current
state-of-the-art machine learning systems sometimes achieve good
performance on a specific and narrow task\ldots{} Instead, humans are
able to understand their environment in a more unified way\ldots which
allows them to quickly generalize\ldots on a new task, thanks to their
ability to reuse previously acquired knowledge'' (p.~2).

In contrast, we uncover evidence of AIs' potential to augment or even
replace research assistants and postdoctoral fellows. These positions
constitute steps on the academic progression pathway towards becoming
faculty. If the need for them is obviated then that trajectory may be
disrupted. In addition, we may expect AI to demonstrate analogous
reasoning competence in corporate settings; even positions requiring
nuanced understanding and creative application within complex fields
could be vulnerabile to technological substitution.

\hypertarget{roadmap}{%
\paragraph{Roadmap}\label{roadmap}}

The next section describes our data. Then we describe our methodology
and results. The final section situates our findings and discusses
implications.

\hypertarget{data}{%
\section{Data}\label{data}}

We base our research on 589 articles published over a span of 28 months,
from October 2021 to January 2024, across four leading psychology
journals: \emph{Cognitive Psychology} (62 articles), \emph{Journal of
Experimental Psychology: General} (167 articles), \emph{Journal of
Personality and Social Psychology} (81 articles), and
\emph{Psychological Science} (279 articles). This selection was chosen
to span the period following the last update to the AI's training data
in September 2021. Significantly, over 90\% of these articles (536 out
of 589) were published in 2022 or later, ensuring they appeared at least
three months after the AI's training data cutoff. This temporal scoping
facilitates a comprehensive evaluation of the AI's capacity to analyze
and interpret fresh research content it has not previously encountered.

The dataset displays considerable diversity across multiple dimensions.
From an authorship perspective, it features contributions from 2,237
unique scientists, with 125 authors appearing in two articles, 33 in
three articles, and 4 in four or more articles. The abstracts present in
a wide range of lengths, from 630 to 2,364 characters. The median
character count is 1,166, with an average of 1,308 characters and a
standard deviation of 314 characters. This range not only mirrors the
extensive scope of research within psychology but also lays a solid
foundation for assessing the AI's performance across a diverse set of
scientific studies.

\begin{figure}[htbp]
\centering
\begin{subfigure}{.5\textwidth}
  \centering
  \includegraphics[width=\linewidth]{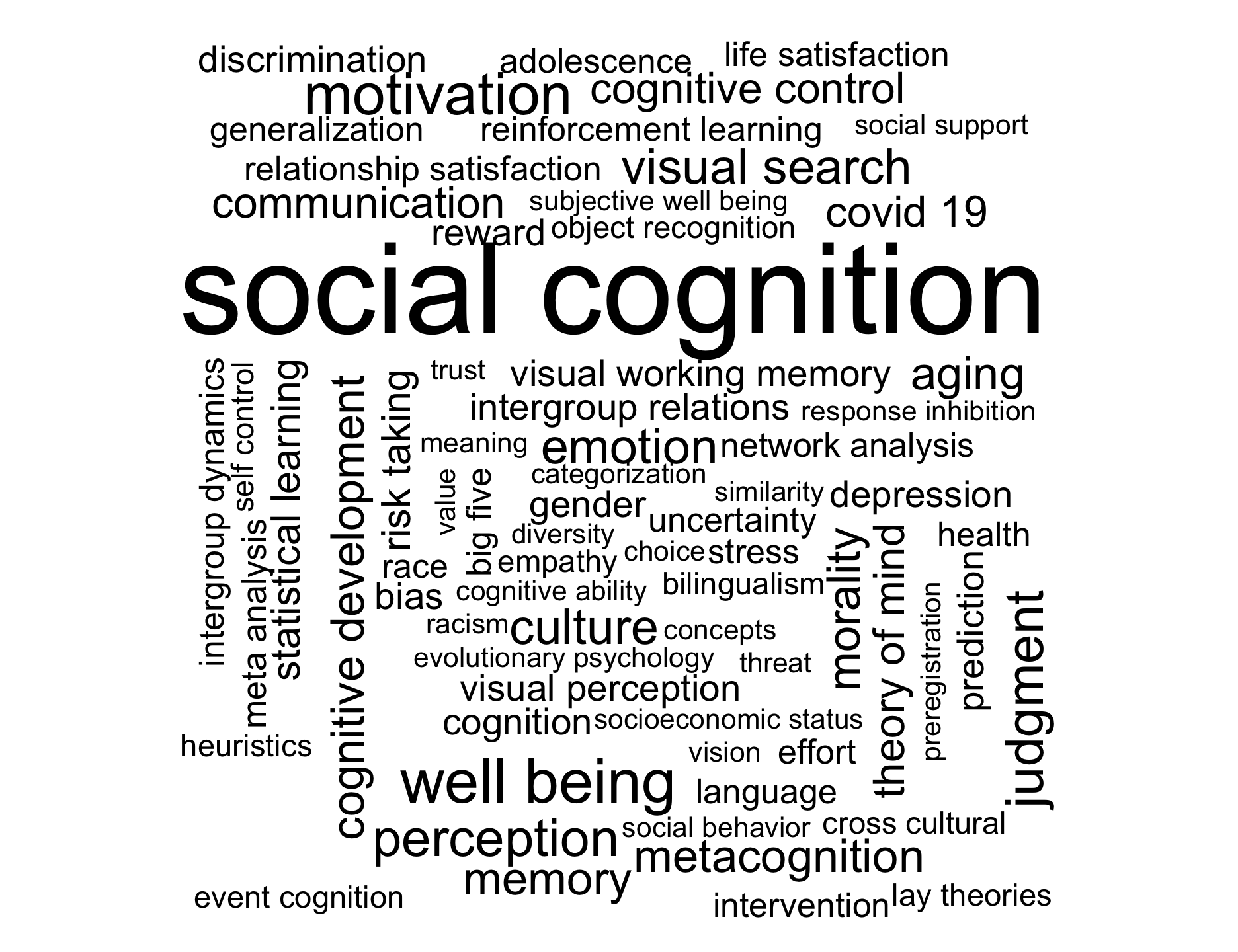}
  \caption{Word Cloud of Author-Specified Keywords}
\end{subfigure}%
\begin{subfigure}{.5\textwidth}
  \centering
  \includegraphics[width=\linewidth]{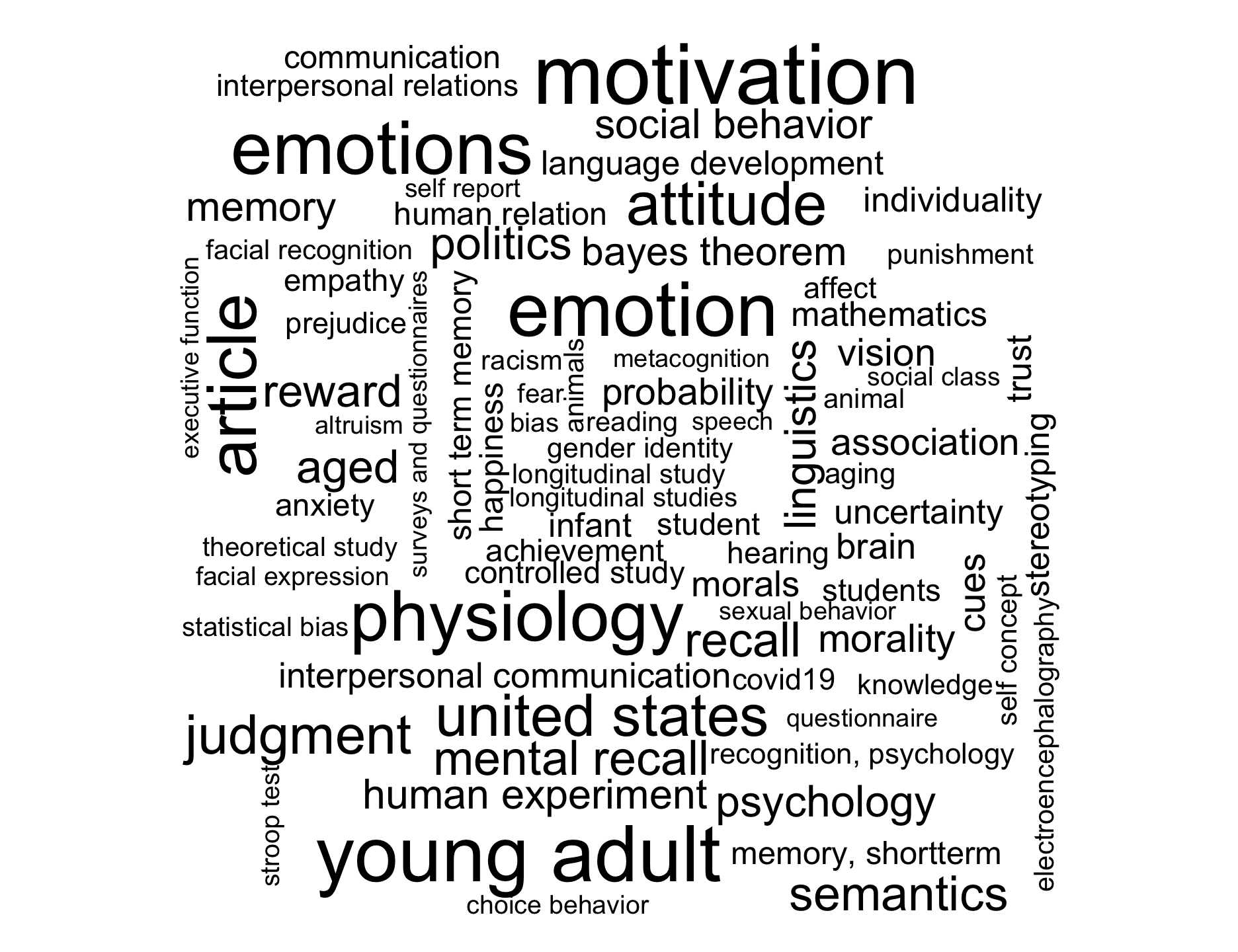}
  \caption{Word Cloud of Indexed Keywords}
\end{subfigure}
\caption{Word Clouds of Author-Specified and Indexed Keywords}
\label{fig:wordclouds_abstracts}
\begin{minipage}{\linewidth}
\medskip
\footnotesize
Note: The word clouds are derived from both author-specified and indexed keywords, following the exclusion of uninformative or ubiquitous terms (e.g., `study,' `results') to emphasize the dataset's thematic focus. Author-specified keywords are chosen by the contributors, while indexed keywords are subject headings from Scopus' thesaurus, facilitating standardized cross-study comparisons.
\end{minipage}
\end{figure}

To evaluate the dataset's thematic diversity, Figure
\ref{fig:wordclouds_abstracts} showcases word clouds derived from both
author-specified and indexed keywords. The author-specified keywords,
selected by the contributors themselves, illuminate the core themes of
their research. Conversely, indexed keywords, sourced from Scopus'
thesaurus, offer a standardized linguistic framework that aids in
thorough cross-study comparisons. Common terms in psychological
literature, such as `open data,' `open materials,' and various
demographic descriptors, were intentionally omitted to accentuate the
dataset's thematic concentration.

The author-specified keywords exhibit a rich variety of research
interests, with the most prevalent terms being social cognition (36
occurrences), decision-making (35), attention (15), well-being (15),
motivation (14), working memory (14), individual differences (12),
judgment (12), perception (12), and culture (11). This array underscores
the areas of significant contemporary academic focus. On the other hand,
the indexed keywords highlight the concepts most frequently explored,
featuring learning (142 occurrences), cognition (120), child (108),
decision-making (103), motivation (88), young adult (80), attention
(78), emotion (77), emotions (145), and article (68). Together, these
sets of keywords underscore the dataset's extensive reach across a
multitude of psychological subfields.

\hypertarget{methodology-and-results}{%
\section{Methodology and Results}\label{methodology-and-results}}

Our approach comprises five distinct steps:

\begin{enumerate}
\def\labelenumi{\arabic{enumi}.}
\item
  \textbf{Stimulus Construction}: The first step refines a scientific
  abstract into a redacted form. The goal here is twofold: to remove any
  direct mention of empirical results while safeguarding the abstract's
  original intent and inquiry. This process assesses the AI's skill in
  navigating and restructuring sophisticated scientific discourse, while
  maintaining the integrity of the research narrative. Success in this
  task hinges on the AI's linguistic dexterity and its comprehensive
  understanding of the topic at hand.
\item
  \textbf{Redaction Assessment}: The second step assesses the extent to
  which the redacted research conceals empirical findings without
  distorting the original research context or questions. Utilizing a
  comprehensive nine-point rubric, this step highlights the AI's ability
  for in-depth analysis and discerning judgment, ensuring the prior
  redaction meets stringent criteria for both concealment and accuracy.
\item
  \textbf{Prediction}: The third step challenges an AI instance to
  generate specific, quantifiable predictions based on the redacted
  variant. It simulates a scenario where the AI must infer study
  outcomes and theoretical implications without direct access to
  empirical data. Thus, it tests the AI's ability to engage in deductive
  reasoning and theoretical extrapolation based on limited information,
  mirroring the cognitive processes researchers engage in when
  formulating hypotheses based on existing literature.
\item
  \textbf{Prediction Assessment}: The fourth step compares between the
  AI-generated predictions and the actual findings and implications
  reported in the original research. This assessment not only evaluates
  the accuracy of the AI's predictions but also examines the depth of
  the AI's understanding of the research. It involves both qualitative
  and quantitative analysis, identifying patterns, errors, or
  limitations in the AI's reasoning, and providing reasoning and
  illustrative examples to substantiate its findings.
\item
  \textbf{Rubric-Based Evaluation}: The fifth step quantifies the
  analysis in Step 4 using a detailed rubric. The rubric assesses AI
  performance on empirical outcomes and theoretical implications,
  providing a standardized measure of its predictive accuracy and
  conceptual understanding. This structured evaluation measures the
  extent to which the AI's predictions align with human expert reasoning
  and findings in scientific research.
\end{enumerate}

We employ separate AI instances for each step, with no overlap in data
or memory, enabling an accurate assessment of each task in isolation.
This method reflects the compartmentalization seen in human cognitive
processes, making our results more comparable to human expertise, and
enhancing the interpretability of our design and findings.

In certain steps, we employ an AI to critique the work of other AIs to
mitigate potential shortcomings of human workers who may overlook
details or unintentional disclosures. The rationale behind this approach
is rooted in the understanding that an AI of the same model would be
most adept at recognizing signals or information hidden within redacted
text, given its familiarity with data patterns and its own processing
logic, that might otherwise aid an AI to cheat on this test. Therefore,
to ensure true redaction, minimize information leakage, and ensure
fairness we feature the same or similar AI model for all tasks.

Moreover, we enforce a strict separation between the AI instances
engaged in different stages of the experiment to ensure the assessment
process remains unbiased and uncontaminated by prior knowledge, closely
simulating the conditions under which human experts operate when
evaluating new research, such as during peer review
(\protect\hyperlink{ref-goeken2020rise}{Goeken et al. 2020}). This
methodological choice not only preserves the integrity and validity of
the assessment process but also strengthens the relevance and
authenticity of our study, enhancing its applicability to real-world
scenarios.

To facilitate interactions between the research infrastructure and the
AI instances, we utilize the official OpenAI Application Programming
Interface (API), specifically employing the `gpt-4-0613' model with a
temperature setting of 0. The API serves two primary functions: it sends
step-specific prompts to the respective instances and retrieves the
generated text for further analysis. Utilizing the API offers three
significant benefits: it automates methodological tasks, eliminating
manual intervention and reducing the risk of human error; it provides
flexibility to adjust the model's parameters and prompts for each
research task; and it systematically records all interactions, enhancing
research transparency and facilitating future replication efforts. While
our current study leverages automated API interactions, our methodology
could also be implemented manually through an interface that provides
interactive access to a suitable AI.

A consequence of our careful and conservative methodology is that it
incurs time and monetary costs that limit the scale of journals and
articles we can study. This is because each article and each step
requires the instantiation of a new instance from the base model, and
therefore does not enable parallel processing. The study cannot be
conducted in parallel, whereby all articles are sent in a single API
call in each step, necessitating careful and sequential processing;
running the study as it currently stands requires more than 36 hours of
constant API calls.

As our purpose is primarily to provide detailed and statistically sound
evidence, we chose to focus on four journals---the three biggest
journals in psychology and a relatively less prestigious journal as a
contrast. With developments and improvements, and a reduction in API
costs, we expect that these computational considerations will mitigate
automatically, enabling broader analysis. In this paper, we focus on the
theoretical considerations to establish prima facie grounds for
anticipating the extent of displacement AI may cause.

Finally, a limitation intrinsic to the subject matter we study is that
our findings are deeply nuanced, as they reflect predictions and
analyses of published research in eminent psychological journals; there
is simply no straightforward way to convert research predictions,
assessments of accuracy, and other such measures into numeric metrics
for quantitative analysis. We approach this matter in three ways. First,
we provide a running example, where, in each major step, we showcase
each step in detail. This allows for nuanced qualitative exploration.
Second, we provide abstractive summaries where applicable, pooling
qualitative information across studies to identify key commonalities.
Third, we develop and apply rubrics to convert qualitative insights into
quantitative assessments, facilitating statistical analysis. Jointly,
across these approaches, we seek to provide a detailed and comprehensive
view of the AI's efficacy.

\hypertarget{step-1-stimulus-construction}{%
\subsection{Step 1: Stimulus
Construction}\label{step-1-stimulus-construction}}

The initial step involves presenting an AI with by a predefined prompt,
along with the original, unredacted research. The prompt is structured
to direct the instance to generate redacted research that preserves the
integrity and coherence of the original research narrative without
inadvertently revealing empirical data or conclusions.\footnote{To cater
  to the interpretive capabilities of AI instances, we employ prompts
  that are significantly more detailed than those typically used in
  human-oriented studies. This approach aligns with the nuanced
  processing abilities of AIs, leveraging their analytical strengths. It
  also provides sufficient detail for the instances to implement the
  prompts programmatically and independently, ensuring clarity and
  reducing ambiguity.}

\singlespacing
\begin{quoting}
\noindent You are tasked with editing a scientific abstract to create a redacted version. Your primary objective is to maintain the research context and questions while scrupulously obfuscating any empirical findings. Begin by identifying elements—sentences, phrases, data, statistics—that explicitly or implicitly convey empirical outcomes. Subsequent to identification, either excise these elements or replace them with abstract placeholders or indeterminate language; however, this action should not distort the original research context or questions. Uphold the integrity of the research context and questions, either preserving them in their original formulation or rephrasing them in a manner that retains their essential meaning. Exercise vigilant caution to prevent the disclosure of empirical findings, particularly by avoiding descriptors that signal the magnitude or direction of effects. Preserve details pertaining to the study sample, experimental design, and methodologies to maintain context. Your meticulous adherence to these guidelines is pivotal for safeguarding the methodological integrity of subsequent steps in this research study.
\end{quoting}
\doublespacing

Table \ref{table:comparison} revisits the research discussed in the
introduction of our paper. To facilitate analysis, we divided it into
four foundational thematic components: Introduction, Methods, Results,
and Conclusion. This division into components was not provided to the
AI, which received the original research in its published form. The
table juxtaposes the original and redacted research, illustrating the
challenge of selectively excising empirical information while preserving
a coherent narrative.

\begin{table}[htbp]
\begin{tabularx}{\textwidth}{|X|}
\hline
\multicolumn{1}{|c|}{\textbf{Original}} \\
\hline
\textbf{Introduction:} Not all older adults with dementia-related neuropathology in their brains experience cognitive decline or impairment. Instead, some people maintain relatively normal cognitive functioning despite neuropathologic burden, a phenomenon called cognitive resilience. \\
\textbf{Methods:} Using a longitudinal, epidemiological, clinical-pathologic cohort study of older adults in the United States (N = 348), the present research investigated associations between well-being and cognitive resilience. \\
\textbf{Results:} Consistent with preregistered hypotheses, results showed that higher eudaimonic well-being (measured via the Ryff Psychological Well-Being Scale) and higher hedonic well-being (measured via the Satisfaction with Life Scale) were associated with better-than-expected cognitive functioning relative to one’s neuropathological burden (i.e., beta-amyloid, neurofibrillary tangles, Lewy bodies, vascular pathologies, hippocampal sclerosis, and TDP-43). \\
\textbf{Conclusion:} This research highlights the importance of considering eudaimonic well-being in efforts to prevent dementia. \\
\hline
\multicolumn{1}{|c|}{\textbf{Redacted}} \\
\hline
\textbf{Introduction:} Not all older adults with dementia-related neuropathology in their brains experience cognitive decline or impairment. Instead, some people maintain relatively normal cognitive functioning despite neuropathologic burden, a phenomenon called cognitive resilience. \\
\textbf{Methods:} Using a longitudinal, epidemiological, clinical-pathologic cohort study of older adults in the United States (N = 348), the current research examined possible associations between well-being and cognitive resilience. \\
\textbf{Results:} The study considered both eudaimonic well-being (measured via the Ryff Psychological Well-Being Scale) and hedonic well-being (measured via the Satisfaction with Life Scale) in its investigation of cognitive functioning in relation to one’s neuropathological burden (namely, beta-amyloid, neurofibrillary tangles, Lewy bodies, vascular pathologies, hippocampal sclerosis, and TDP-43). \\
\textbf{Conclusion:} The potential implications of these explorations underscore the importance of considering various dimensions of well-being in efforts to prevent dementia. \\
\hline
\end{tabularx}
\caption{Comparison of Original and Redacted Research}
\label{table:comparison}
\begin{minipage}{\linewidth}
\medskip
\footnotesize
Note: The table contrasts the original and redacted research, segmented into thematic components: Introduction, Methods, Results, and Conclusion. The redacted variants are carefully edited versions of the originals, with empirical findings either removed or obscured to prevent direct disclosure, while still preserving the overarching research context.
\end{minipage}
\end{table}

A close examination reveals that the integrity of the introduction and
methods sections is carefully preserved across both the original and
redacted versions. This ensures that the foundational context and
investigative framework of the study remain intact, allowing for a clear
understanding of the research's scope and objectives without any
dilution of its methodological rigor. The original version, however,
discloses a significant empirical finding, stating that `higher
eudaimonic well-being\ldots was associated with better-than-expected
cognitive functioning.' This explicit revelation of results directly
communicates the study's outcomes, providing concrete evidence of the
research's hypotheses being met.

Contrastingly, the redacted version adopts a more guarded approach in
discussing the results and conclusions. By suggesting `possible
associations' and hinting at `potential implications,' it strategically
avoids revealing specific empirical outcomes. This nuanced redaction
effectively maintains the narrative's focus on `cognitive resilience'
without compromising the research's conceptual framework. The
transformation from explicit empirical disclosure to a more abstract
discussion not only obfuscates the direct findings but also shifts the
emphasis towards the broader theoretical implications of the study. This
careful balancing act ensures that the essence of the research inquiry
is communicated while safeguarding the specifics of the empirical
evidence, thereby preserving the narrative's integrity and fostering an
environment conducive to speculative inquiry based on the provided
context.

To systematically analyze the redaction process, we incorporate topic
modeling as an analytical tool. Topic models are statistical frameworks
designed to identify latent topics within a corpus by categorizing
documents according to their word distribution patterns
(\protect\hyperlink{ref-park2015automatic}{Park et al. 2015}).
Specifically, we apply a state-of-the-art topic modeling approach called
BERTopic (\protect\hyperlink{ref-grootendorst2022bertopic}{Grootendorst
2022}), which utilizes transformer-based language representations to
achieve heightened sensitivity compared to conventional topic models.

The utilization of topic modeling in our analysis serves a dual purpose.
Firstly, it verifies that the thematic essence of the introduction and
methods sections remains unchanged between the original and redacted
versions. This is crucial for maintaining the research narrative's
coherence. Secondly, it facilitates a focused evaluation of the
redaction's effectiveness in obfuscating the specific empirical findings
within the results and conclusion sections. This approach allows us to
systematically scrutinize the redacted text, ensuring that the
introduction and methods retain their thematic focus and integrity,
thereby supporting a comprehensive understanding of the research's
objectives and methodologies.

We estimate two distinct topic models: one for the original research and
another for the redacted variants. By estimating separate topic models
for the original and redacted collections, we obtain condensed
representations of their key thematic elements. Comparing these topic
models then enables us to evaluate whether essential contextual features
are preserved during redaction. If the core themes persist between the
original and redacted versions, it suggests effective empirical
obfuscation without loss of narrative integrity.

\begin{table}[htbp]
\centering
\begin{tabularx}{\textwidth}{|X|X|}
\hline
\multicolumn{1}{|c|}{\textbf{Topics from Original Research}} & \multicolumn{1}{|c|}{\textbf{Topics from Redacted Research}} \\
\hline
Visual Working Memory Experiments & Visual Perception and Object Processing \\
Linguistics and Cognitive Processing in Language Learning & Language Development and Semantic Structures in Bilingualism Research \\
Visual Perception and Attention Processing & Visual Perception and Object Processing \\
The Relationship between Social Interactions, Activities, and Well-being Over Time & Relationship between Social Interactions, Personal Satisfaction, and Well-being \\
The Impact of Prosocial Acts and Giving on Both Givers and Recipients & Prosocial Behavior and its Psychological Impact \\
Decision Making and Choice Models & Decision Making Models and Processes \\
Perception and Interpretation of Facial Traits and Racial Stereotypes & Perception and Impression Formation Based on Faces \\
Moral Psychology and Perception of Harm & Moral Judgments and its Influences in Society \\
Climate Beliefs and Political Influence on Scientific Consensus & Racial and Political Attitudes in Intergroup Interactions \\
Personality Traits Development Across Adulthood and Generations & Personality Traits and Adulthood Development in the Big Five Framework \\
Bias in Perception of Future Events and Outcomes & Decision Making and Risk Perception \\
Perception and Impression Formation from Voices and Faces & Perception and Impression Formation Based on Faces \\
Cognitive Decline and Cognitive Resilience with Aging & Cognitive Decline and Intelligence in Aging Adults \\
Adolescent Decision Making: Risk, Reward and Delay Gratification & Decision Making and Risk Perception \\
Racial Prejudice and Militarization in the US & Racial and Political Attitudes in Intergroup Interactions \\
\hline
\end{tabularx}
\caption{Topics from Original and Redacted Research}
\label{table:comparison_topic_models}
\begin{minipage}{\linewidth}
\medskip
\footnotesize
Note: Redacted research are versions of original scientific research edited to remove or conceal empirically revealing elements, while preserving the overall research context and questions. Topics were inferred using the BERTopic model (Grootendorst 2022), which combines transformer models and class-based TF-IDF to generate interpretable topics with meaningful keywords. This model overcomes limitations of earlier models like LDA, offering improved semantic understanding and multilingual document embedding for enhanced contextual representation.
\end{minipage}
\end{table}

Table \ref{table:comparison_topic_models} compares the inferred topics
from the two sets, providing specific examples to assess thematic
coherence. For instance, the topic `Visual Working Memory Experiments'
from the original abstracts and its redacted counterpart `Visual
Perception and Object Processing' demonstrate how core themes, such as
visual processing, are preserved while modifying specific elements like
`working memory' to `object processing.' Similarly, `Linguistics and
Cognitive Processing in Language Learning' aligns with `Language
Development and Semantic Structures in Bilingualism Research' in the
redacted abstracts, retaining the central theme of language learning
while introducing nuanced variations in the context of bilingualism.
This shift not only maintains the essence of the research focus but also
adapts it to encompass a broader scope within the field of linguistics,
demonstrating the AI's ability to preserve thematic relevance while
ensuring empirical obfuscation.

Another notable pairing is `Decision Making and Choice Models' with
`Decision Making Models and Processes.' The fundamental focus on
decision-making processes is maintained, with the redacted version
presenting a broader perspective on the models and processes involved.
In the case of `Perception and Interpretation of Facial Traits and
Racial Stereotypes,' its redacted parallel `Perception and Impression
Formation Based on Faces' successfully conserves the primary focus on
facial perception, subtly altering the context to remove specific
empirical details.

These examples demonstrate that the AI-generated redactions broadly
retain thematic coherence relative to the original abstracts. While
adjustments are introduced to excise empirical details, the core
research foci persist, spanning domains from visual and linguistic
processing to decision-making and facial perception. This selective
empirical obfuscation crucially upholds narrative integrity. More
broadly, these redaction examples highlight modern AI's capacity for
nuanced textual alterations that balance competing objectives---removing
empirical data while preserving conceptual meaning. This ability likely
stems from the model's architectural depth, which supports sensitive
semantic manipulation. Overall, the observed high efficacy affirms the
promise of this methodology for generating controlled experimental
stimuli to rigorously test AI reasoning faculties.

\hypertarget{step-2-redaction-assessment}{%
\subsection{Step 2: Redaction
Assessment}\label{step-2-redaction-assessment}}

Having established topic coherence, thereby ensuring systematic
conceptual consistency between the original and redacted abstracts in
terms of introduction and focus, we employ a second AI instance for the
evaluation of redacted abstracts. This step focuses on two dimensions:
(1) the preservation of contextual integrity within each abstract pair,
and (2) the successful concealment of empirical findings in the redacted
versions. Operating independently from the AI used in Step 1, this
instance is tasked with conducting a detailed comparison between the
original and redacted abstract pairs. It has no prior knowledge of the
redacted abstract's origins, thereby reducing the risk of inadvertent
information leakage. The following prompt guides its evaluation:

\singlespacing
\begin{quoting}
\noindent You are tasked with conducting a comparative assessment of an original and a redacted scientific abstract. No prior knowledge of the study's methodology or redaction guidelines should influence your evaluation. Begin your qualitative analysis by juxtaposing the original and redacted abstracts, focusing on their comparative elements. Assess the efficacy of the redacted version in concealing empirical outcomes, while being vigilant for any traces that could inadvertently disclose the study's findings. Evaluate the success of the redacted abstract in preserving the integrity of the original research context and questions, regardless of whether the text is reproduced verbatim or effectively paraphrased. Conclude your assessment in a clearly delineated section by assigning a numerical rating to the redaction's efficacy using a nine-point scale: 1 indicates 'ineffective concealment coupled with contextual distortion,' while 9 signifies 'exemplary efficacy in obfuscating empirical findings while preserving contextual integrity.' Format this numerical rating as a structured string: 'Quantitative Rating: [Your numerical rating here]'. Your adherence to this structured output format is pivotal for ensuring the output can be programmatically parsed and for safeguarding the methodological integrity of subsequent research steps.
\end{quoting}
\doublespacing

This rubric is designed for an instance with the same model architecture
and knowledge representation as the redacting AI but uninformed of the
study's purpose and any intermediate representations from the prior
instance, creating a consistent and robust framework for assessment:
Uniformity across the ontological and epistemological foundations of the
AI, stemming from identical model structure and training, enables the
evaluative instance to precisely gauge the redaction efficacy in hiding
empirical findings while maintaining research context integrity.
Moreover, ensuring each instance remains uninformed of the study's
overarching goals or the specifics of preceding steps preserves the
research design's validity and guards against potential biases that
could affect the evaluation process. The rubric is applied through
distinct and independent API calls, ensuring that for each abstract
pair, the evaluating instance is only aware of the prompt, the rubric,
and the abstract pair itself.

\begin{figure}[htbp]
\centering
\includegraphics[width=0.75\textwidth]{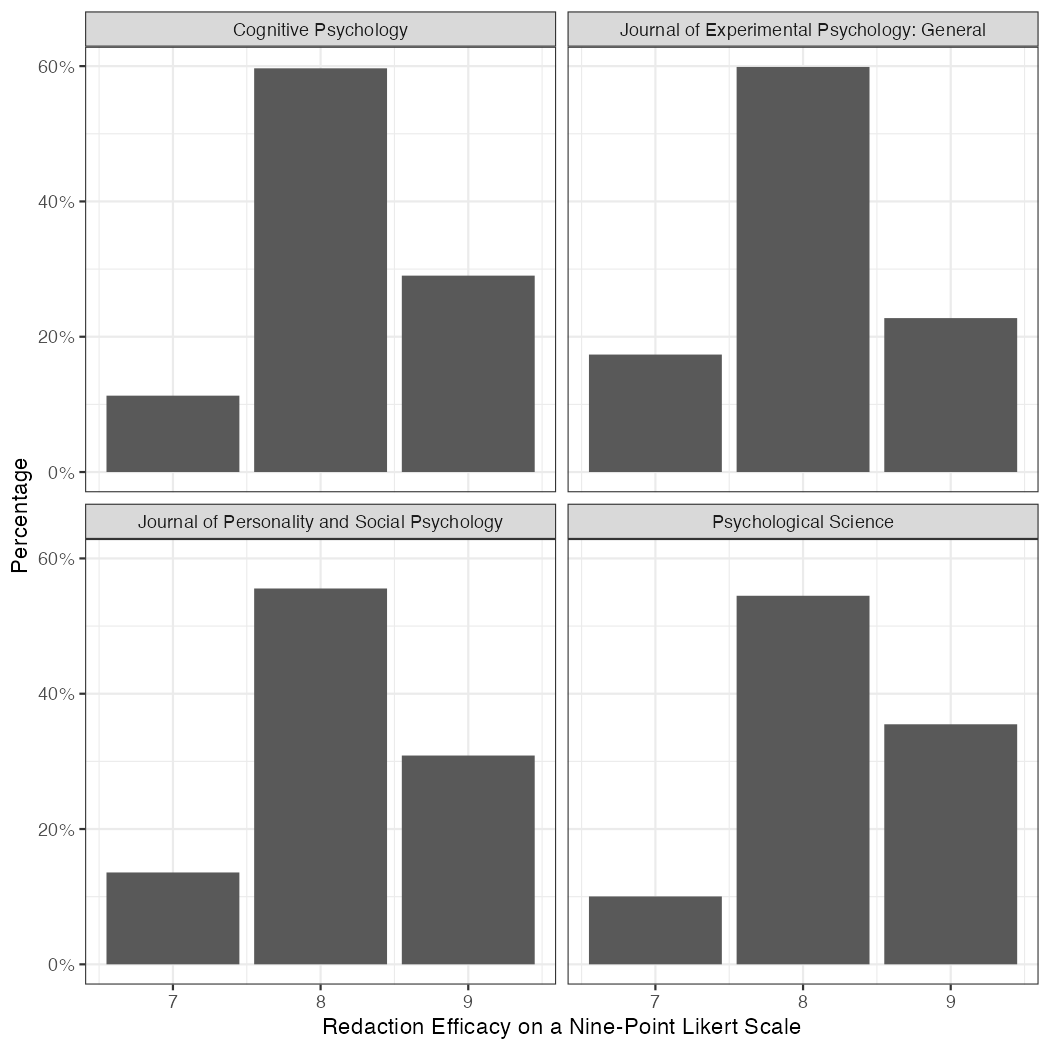}
\caption{Bar Plot of Redaction Efficacy Ratings}
\label{fig:bar_redaction_efficacy}
\begin{minipage}{\linewidth}
\medskip
\footnotesize
Note: The bar plot presents redaction efficacy ratings, measured on a nine-point scale, as percentages for each academic journal.
\end{minipage}
\end{figure}

The distribution of redaction efficacy ratings, as presented in Table
\ref{table:redaction_efficacy}, offers intriguing insights into the
nuanced performance of AI in redacting scientific abstracts across
various academic journals. Notably, the ratings predominantly cluster
around a central tendency of 8, indicating a generally high level of
redaction efficacy across the board. Figure
\ref{fig:bar_redaction_efficacy} illustrates the distribution of
redaction efficacy ratings, faceted by academic journal. A closer
examination of both the figure and the table reveals subtle variations
in performance that merit further discussion.

For instance, \emph{Cognitive Psychology} exhibits a relatively balanced
distribution of ratings, with a significant majority (59.7\%) receiving
a rating of 8, and a noteworthy proportion (29.0\%) achieving the
highest rating of 9. This suggests that the AI's redaction performance
is particularly effective in the context of cognitive psychology
research, possibly due to the specific nature of the content or the
clarity of the abstracts in this journal. Conversely, \emph{Journal of
Experimental Psychology: General} shows a slightly lower proportion of
the highest ratings (22.8\% receiving a 9), despite a similar majority
(59.9\%) receiving an 8. This could imply that the broader scope of
topics covered by this journal introduces complexities that slightly
challenge the AI's ability to redact without compromising the integrity
of the research narrative.

\emph{Journal of Personality and Social Psychology} and
\emph{Psychological Science} both demonstrate strong redaction efficacy,
with over 30\% of abstracts in each journal receiving the highest rating
of 9. This high level of performance underscores the AI's capability to
handle a wide range of topics within psychology, from the intricacies of
personality research to the diverse studies published in
\emph{Psychological Science}. The slight variations in redaction
efficacy ratings across journals may reflect differences in the
complexity of topics, the specificity of language used in abstracts, or
the inherent challenges associated with accurately redacting empirical
findings while preserving the original research context. These findings
highlight the AI's proficiency in navigating and restructuring
sophisticated scientific discourse across a spectrum of psychological
research areas.

Moreover, the consistent achievement of high ratings across all journals
underscores the potential of AI as a valuable tool in the scientific
research process, capable of performing tasks that require a deep
understanding of content and context. The ability of AI to maintain the
integrity of the research narrative while effectively concealing
empirical outcomes is particularly promising for applications in peer
review processes, research synthesis, and the development of educational
materials.

\begin{table}[htbp]
\centering
\caption{Distribution of Redaction Efficacy Ratings by Academic Journal}
\label{table:redaction_efficacy}
\begin{tabularx}{\textwidth}{|X|c|c|c|}
\hline
\textbf{Journal} & \textbf{Rating} & \textbf{Count} & \textbf{Percentage (\%)} \\
\hline
Cognitive Psychology & 7 & 7 & 11.3 \\
Cognitive Psychology & 8 & 37 & 59.7 \\
Cognitive Psychology & 9 & 18 & 29.0 \\
Journal of Experimental Psychology: General & 7 & 29 & 17.4 \\
Journal of Experimental Psychology: General & 8 & 100 & 59.9 \\
Journal of Experimental Psychology: General & 9 & 38 & 22.8 \\
Journal of Personality and Social Psychology & 7 & 11 & 13.6 \\
Journal of Personality and Social Psychology & 8 & 45 & 55.6 \\
Journal of Personality and Social Psychology & 9 & 25 & 30.9 \\
Psychological Science & 7 & 28 & 10.0 \\
Psychological Science & 8 & 152 & 54.5 \\
Psychological Science & 9 & 99 & 35.5 \\
\hline
\end{tabularx}
\begin{minipage}{\textwidth}
\vspace{0.5cm}
\footnotesize
Note: This table presents the distribution of redaction efficacy ratings, measured on a nine-point scale, across four academic journals. The ratings are categorized by their numeric value (7 to 9), with the count and percentage of each rating provided for each journal.
\end{minipage}
\end{table}

\begin{table}[htbp]
\centering
\caption{P-values of Variations in Redaction Efficacy Ratings Across Academic Journals}
\label{table:redaction_efficacy_pvalues}
\begin{tabularx}{\textwidth}{|X|c|c|c|}
  \hline
 & \textbf{JEP:G} & \textbf{JPSP} & \textbf{PS} \\ 
  \hline
  Differences in Mean Ratings: & & & \\
  \hspace{0.5cm} Cognitive Psychology & 0.18 & 0.97 & 0.38 \\ 
  \hspace{0.5cm} Journal of Experimental Psychology: General (JEP:G) &  & 0.17 & 0.00 \\ 
  \hspace{0.5cm} Journal of Personality and Social Psychology (JPSP) &  &  & 0.32 \\ 
  Differences in Variance Ratings: & & & \\
  \hspace{0.5cm} Cognitive Psychology & 0.98 & 0.62 & 0.46 \\ 
  \hspace{0.5cm} Journal of Experimental Psychology: General (JEP:G) &  & 0.52 & 0.27 \\ 
  \hspace{0.5cm} Journal of Personality and Social Psychology (JPSP) &  &  & 0.86 \\ 
  \hline
\end{tabularx}
\begin{minipage}{\textwidth}
\vspace{0.5cm}
\footnotesize
Note: This table presents the p-values of differences in redaction efficacy ratings, measured on a nine-point scale, across four academic journals. JEP:G = {\it Journal of Experimental Psychology: General}; JPSP = {\it Journal of Personality and Social Psychology}; and PS = {\it Psychological Science}. The p-values are derived from t-tests (for differences in means) and Levene's tests (for differences in variances), testing pairwise across all journals, to assess the statistical significance of variations in redaction efficacy.
\end{minipage}
\end{table}

To further explore the consistency of redaction efficacy across
journals, we present the p-values from t-tests and Levene's tests in
Table \ref{table:redaction_efficacy_pvalues}. The t-test assesses
differences in the average ratings between journals, while the Levene
test evaluates if the variability in ratings differs significantly.
Together, these tests examine the variability in redaction efficacy
across academic contexts.

We find that, although there are some variations in the means and
variances of redaction efficacy ratings, these differences are not
consistently statistically significant. This suggests a generally
uniform efficacy of the AI in redacting abstracts across various
journals in psychology, and therefore, across creative content with
varying complexity and specificity of language, as well as differing
influences of novelty and conceptual knowledge. Thus, these findings
underscore the AI's robustness and adaptability in handling diverse and
emergent content, reinforcing its potential as a valuable aid for
fostering creativity in complex and dynamic domains.

\hypertarget{step-3-prediction}{%
\subsection{Step 3: Prediction}\label{step-3-prediction}}

A third instance is presented solely with the redacted research, without
access to the original. The objective of this third step is to evaluate
the AI's analytical rigor in using its knowledge of human behavior and
prior theory to make empirically grounded predictions based on the
redacted research. Thus, the task probes the depth of the AI's knowledge
and its capacity for inference under information constraints.

It's important to note that the third step employs knowledge learned
through two mechanisms. First, knowledge may be gained through
\emph{implicit learning}, which occurs without explicit intention,
through observation or exposure rather than direct instruction. This
form of learning enables AI to assimilate complex environmental
structures, akin to how one might learn a language through immersion or
understand social norms by observing group interactions. For AI, this
learning is facilitated by analyzing human interactions within its
training data, such as forum posts, and through reinforcement learning
from human feedback.

Second, knowledge may be gained through \emph{explicit learning}, which
involves the deliberate acquisition of knowledge through formal
education and instruction. This method is characterized by the direct
teaching of rules, principles, or concepts. For AI, explicit learning
might involve training on specific datasets, such as preparing for an AP
psychology exam, where the focus is on memorizing material and
understanding formal theories.

These dual modes are complementary in this task. On the one hand,
implicit learning combined with inductive reasoning may facilitate the
AI's prediction of empirical outcomes based on theory, given the vast
data on human interactions that it has been exposed to. On the other
hand, explicit learning combined with deductive reasoning may enable it
to forecast based on what it knows about the formal concepts and
theories in human psychology. These two mechanisms, jointly, provide the
central impetus in the prediction task set out in the following prompt,
used to guide the instance:

\singlespacing
\begin{quoting}
\noindent You are tasked with generating specific and quantifiable predictions based on a redacted scientific abstract. Begin by conducting a systematic analysis of the research context, questions, methods, and procedures detailed in the redacted abstract. Subsequently, formulate explicit and unambiguous empirical predictions that provide quantifiable expectations concerning the study's outcomes. Extend these predictions to encompass theoretical implications, contemplating how the expected outcomes may corroborate, challenge, or necessitate the modification of existing theoretical frameworks. Note any limitations or constraints, such as potential biases or issues with generalizability, that could influence the interpretation of your predictions. Given that the abstract is redacted, your focus should be on the formation of empirically substantiated predictions within the bounds of the available information. Your analytical rigor is essential for enabling a subsequent, rigorous comparative evaluation against the actual empirical findings.
\end{quoting}
\doublespacing

Table \ref{table:prediction} presents the original research alongside
the predictions generated by the AI for our running example, which
investigates the relationship between well-being and cognitive
resilience in older adults, taking into account the impact of
neuropathological burden. This study is used here to demonstrate the
performance of the AI in a specific case study; steps 4 and 5 of the
methodology provide more systematic analyses of the AI's performance
across all 589 studies in our data.

The AI's predictions, derived from a redacted abstract, accurately
forecast a positive correlation between eudaimonic and hedonic
well-being and cognitive functioning, showcasing its adeptness at
sophisticated deductive reasoning. Notably, the AI extends its
predictions to quantifiable measures, suggesting the use of correlation
coefficients and multivariate regression analyses to test these
hypotheses. This demonstrates the AI's capability to engage in detailed,
data-driven analysis.

\begin{table}[htbp]
\begin{tabularx}{\textwidth}{|X|}
\hline
\multicolumn{1}{|c|}{\textbf{Original Abstract}} \\
\hline
Not all older adults with dementia-related neuropathology in their brains experience cognitive decline or impairment. Instead, some people maintain relatively normal cognitive functioning despite neuropathologic burden, a phenomenon called cognitive resilience. Using a longitudinal, epidemiological, clinical-pathologic cohort study of older adults in the United States (N = 348), the present research investigated associations between well-being and cognitive resilience. Consistent with preregistered hypotheses, results showed that higher eudaimonic well-being (measured via the Ryff Psychological Well-Being Scale) and higher hedonic well-being (measured via the Satisfaction with Life Scale) were associated with better-than-expected cognitive functioning relative to one’s neuropathological burden (i.e., beta-amyloid, neurofibrillary tangles, Lewy bodies, vascular pathologies, hippocampal sclerosis, and TDP-43). This research highlights the importance of considering eudaimonic well-being in efforts to prevent dementia. \\
\hline
\multicolumn{1}{|c|}{\textbf{Predictions from Redacted Abstract}} \\
\hline
Based on the provided abstract, several specific and quantifiable predictions can be made:\\
1. Higher levels of eudaimonic well-being and hedonic well-being will be associated with better cognitive functioning relative to neuropathological burden. This could be quantified by a positive correlation between well-being scores and cognitive performance measures, controlling for neuropathological burden.\\
2. The association between well-being and cognitive resilience will remain significant even after controlling for known cognitive resilience factors and dementia risk factors. This could be quantified by a multivariate regression analysis showing that well-being scores significantly predict cognitive performance, even when other factors are included in the model.\\
3. The strength of the association between well-being and cognitive resilience may vary depending on the specific type of neuropathology. This could be quantified by interaction effects in the regression model, showing that the relationship between well-being and cognitive performance is stronger for certain neuropathologies than for others.\\
Theoretical implications of these predictions could include a need to revise existing models of cognitive resilience to incorporate well-being as a key factor. This could challenge the traditional focus on structural and biological factors, suggesting that psychological factors also play a crucial role.\\
However, there are several limitations and constraints to consider. First, the study's cross-sectional design limits the ability to infer causality. It is possible that cognitive resilience leads to higher well-being, rather than the other way around. Second, the study's sample is limited to older adults in the United States, which may limit the generalizability of the findings to other populations. Third, the study relies on self-report measures of well-being, which could be subject to response bias. Finally, the study does not consider potential confounding factors such as medication use or other lifestyle factors that could influence both well-being and cognitive resilience. \\
\hline
\end{tabularx}
\caption{Comparison of Original Abstract and Predictions from Redacted Abstract}
\label{table:prediction}
\begin{minipage}{\linewidth}
\medskip
\footnotesize
Note: The prediction was made on the basis of the original abstract edited to remove empirical findings, while preserving research context and questions.
\end{minipage}
\end{table}

Furthermore, the AI proposes significant theoretical implications,
advocating for a revision of existing cognitive resilience models to
incorporate well-being. This challenges the traditional emphasis on
structural and biological factors, highlighting the AI's potential to
contribute to theoretical advancements in psychology. The AI's ability
to identify critical limitations of the study, including its
cross-sectional design and limited generalizability, further underscores
its capacity for critical evaluation, an essential skill in scientific
inquiry.

The AI's hypothesis regarding the variability in the association between
well-being and cognitive resilience based on different neuropathologies
indicates a nuanced understanding of complex interactions within
psychological research. This insight, along with the AI's comprehensive
analysis, underscores its sophisticated deductive reasoning capabilities
and its potential as a valuable tool for scientific research, despite
the inherent limitations of working with redacted information.

\hypertarget{step-4-prediction-assessment}{%
\subsection{Step 4: Prediction
Assessment}\label{step-4-prediction-assessment}}

In the fourth step, an AI instance evaluates the alignment between
predictions made from redacted research (Step 3) and the published
empirical findings and theoretical implications of the research. This
evaluation, which encompasses both qualitative and quantitative
dimensions, is critical as it informs the systematic rubric-based
analysis of all 589 observations in the subsequent step. For the
remainder of this subsection, we focus on our running example to provide
a detailed qualitative discussion. The next step expands the analysis to
include all observations in our dataset, offering a comprehensive
evaluation of the AI's efficacy as a predictive tool in contemporary
psychology research. These findings are derived by guiding the AI with
the following prompt, presented alongside the original abstract and the
predictions from the prior step:

\singlespacing
\begin{quoting}
\noindent You are presented with an original scientific abstract and a set of predictions based on that abstract. Your mandate is to undertake a comprehensive holistic assessment that scrutinizes the alignment between the predicted empirical findings and theoretical implications, and those reported in the original abstract. Commence by closely examining the predictions, paying special attention to any statistical indicators incorporated within them, and contrast these against the empirical findings and theoretical implications stated in the original abstract. Your assessment should encompass both qualitative and quantitative elements, focusing on the specificity, magnitude, and accuracy of the predictions as they relate to both empirical data and theoretical frameworks. Document any recurring patterns, errors, or limitations observed in the predictions, as such insights are invaluable for understanding the academic utility of language models. Ensure that your holistic assessment is rigorously substantiated through reasoned argumentation and illustrative examples.
\end{quoting}
\doublespacing

\begin{table}[htbp]
\begin{tabularx}{\textwidth}{|X|}
\hline
The predictions align well with the empirical findings and theoretical implications reported in the original abstract. The predictions accurately capture the quantitative elements of the study, such as the positive association between eudaimonic and hedonic well-being and cognitive resilience, and the persistence of this association even after controlling for known cognitive resilience and dementia risk factors. The predictions also correctly anticipate the potential for variation in the strength of the association depending on the specific type of neuropathology.\\
The theoretical implications drawn from the predictions are also consistent with those suggested by the original abstract. Both emphasize the need to consider well-being as a key factor in cognitive resilience, challenging the traditional focus on structural and biological factors and highlighting the role of psychological factors.\\
The predictions also correctly identify several limitations of the study, including the cross-sectional design, the limited generalizability of the findings due to the specific sample, the reliance on self-report measures, and the potential influence of unconsidered confounding factors. These limitations are not explicitly mentioned in the abstract, but they are common issues in this type of research and their inclusion in the predictions demonstrates a nuanced understanding of the research context.\\
However, one prediction that is not directly supported by the abstract is the suggestion that the relationship between well-being and cognitive performance might be stronger for certain neuropathologies than for others. While this is a plausible hypothesis, the abstract does not provide any specific evidence to support it. This prediction seems to be an extrapolation based on the general understanding of the complexity of neuropathological conditions and their interactions with psychological factors, rather than a direct interpretation of the abstract's findings.\\
Overall, the predictions demonstrate a high degree of specificity, magnitude, and accuracy in relation to the empirical data and theoretical frameworks presented in the original abstract. They also highlight potential limitations and constraints of the study, providing a comprehensive and nuanced interpretation of the research. \\
\hline
\end{tabularx}
\caption{Assessment of Predictive Accuracy}
\label{table:prediction_accuracy}
\begin{minipage}{\linewidth}
\medskip
\footnotesize
Note: Predictions were made on the basis of a redacted abstract, which is a version of the original abstract edited to remove or conceal findings, while preserving the research context and questions.
\end{minipage}
\end{table}

Table \ref{table:prediction_accuracy} presents the fourth AI instance's
evaluation of the alignment between the third instance's predictions and
the actual empirical findings and theoretical implications detailed in
the original abstract. This assessment is pivotal, not only for gauging
the AI's capacity to generate informed predictions from limited
information but also for its ability to critically appraise the accuracy
of these predictions.

We now compare the fourth instance's assessment and the actual observed
alignment, as previously discussed and documented in Table
\ref{table:prediction}. The fourth instance demonstrates remarkable
precision in evaluating the congruence between the AI-generated
predictions and the empirical and theoretical content of the original
study. It accurately recognizes the predicted positive correlation
between eudaimonic and hedonic well-being and cognitive resilience,
reflecting the study's primary findings. It also points out that the
original abstract does not explicitly support one of the predictions
regarding the variability in the strength of the association between
well-being and cognitive resilience across different neuropathological
conditions. This prediction suggests the AI's capacity for nuanced
reasoning, enabling it to hypothesize based on a comprehensive
understanding of the interplay between neuropathology and psychological
factors. The fourth instance's detection of this distinction between the
AI prediction and the reported results showcases its ability to
critically appraise the predictions.

Furthermore, the assessment aligns with the theoretical implications
inferred from the predictions, reinforcing the importance of integrating
psychological factors into models of cognitive resilience. This
concordance between the AI's predictions and the original abstract's
insights exemplifies the AI's proficiency in not only grasping but also
contributing to the theoretical discourse in psychology. Additionally,
the fourth instance's acknowledgment of the predictions' identification
of potential research limitations---such as the study's cross-sectional
design and the limited generalizability of its findings---demonstrates
its capacity for critical evaluation. These considerations, though not
explicitly mentioned in the abstract, reflect a deep understanding of
common research challenges.

Next, to ensure a systematic analysis, we provided 100 randomly selected
assessments to a different large-context AI (Claude 2.1) equipped with a
sufficiently broad context window. This AI was given the following
instruction: ``You are presented with 100 assessments of the accuracy of
predictions from redacted abstracts where the empirical findings have
been obfuscated. Please develop an abstractive summary of these
assessments to capture the most frequent and pertinent elements, thereby
providing a consolidated view of the AI's predictive accuracy based on
redacted abstracts.'' We limited the sample to 100 because assessing
more than 100 assessments jointly exceeded the computational
capabilities available with contemporary technology. Additionally, we
used a different AI model that was not informed of the key research
question and findings to ensure the summary was fair and comprehensive.

\begin{table}[htbp]
\begin{tabularx}{\textwidth}{|X|}
\hline
The predictions demonstrated a high degree of alignment with the empirical findings and theoretical implications stated in the original abstracts. In most cases, the predictions accurately anticipated the key results of the studies, including the direction and often the magnitude of effects. The predicted theoretical implications were also largely consistent with those discussed in the abstracts.\\
When discrepancies occurred, they were usually minor and stemmed from a lack of specificity rather than a complete misunderstanding of the study. Common issues included failing to predict a specific result reported in the abstract, making assumptions that extended beyond the scope of the abstract's findings, or lacking concrete statistical indicators to allow comparison with the original data.\\
The predictions often identified relevant limitations and constraints of the studies, even when these details were not explicitly stated in the abstracts. Frequently noted limitations had to do with potential biases, lack of generalizability, reliance on assumptions or specific models, issues with self-reported data, and failure to account for confounding variables.\\
Across all assessments, recurring strengths of the predictions were their overall alignment with reported results, their anticipation of theoretical implications, and their identification of study limitations. Areas needing improvement included greater specificity regarding magnitude of effects, avoiding assumptions not grounded in evidence, and fully accounting for mixed or contrasting results. But in the majority of cases, the predictions demonstrated a strong understanding of the studies and an ability to generate insightful academic analysis regarding expected findings.\\
\hline
\end{tabularx}
\caption{Abstractive Summary of Assessments}
\label{table:summary_accuracy}
\begin{minipage}{\linewidth}
\medskip
\footnotesize
Note: A Claude 2.1 instance, a large context AI model, was used to form the abstractive summary due to its broad context window capability, ensuring a fair and comprehensive synthesis of the 100 assessments. This approach was chosen to provide a consolidated view of the AI's predictive accuracy based on redacted abstracts.
\end{minipage}
\end{table}

The abstractive summary provided by Claude 2.1, as shown in Table
\ref{table:summary_accuracy}, highlights the following key points: The
AI's predictions align closely with the empirical findings and
theoretical implications of the original abstracts, demonstrating a high
degree of accuracy in anticipating study outcomes. While minor
discrepancies occasionally arise due to a lack of specificity or
assumptions extending beyond the abstracts' scope, these instances are
relatively rare. The AI's ability to identify potential study
limitations, even when not explicitly mentioned, underscores its nuanced
understanding of research methodologies and the complexities inherent in
psychological studies. Overall, the predictions reflect a strong grasp
of the subject matter, showcasing the AI's potential as a valuable tool
for academic analysis and insight generation.

\hypertarget{step-5-rubric-based-evaluation}{%
\subsection{Step 5: Rubric-Based
Evaluation}\label{step-5-rubric-based-evaluation}}

In this step, we quantitatively assessed predictive accuracy. We
developed a rubric comprising two components: empirical alignment, which
assesses the closeness of AI predictions to actual research outcomes,
and theoretical alignment, which evaluates the congruence between the
AI's theoretical implications and those in the original research. High
scores in empirical alignment indicate the AI's success in accurately
inferring empirical results, showcasing its deductive reasoning and
prediction capabilities. Similarly, high scores in theoretical alignment
demonstrate the AI's ability to effectively contextualize outcomes
within theoretical frameworks, highlighting its understanding of
scientific discourse. This rubric is incorporated into the following
prompt provided to the AI:

\singlespacing
\begin{quoting}
\noindent You are presented with a holistic assessment that characterizes the accuracy of predictions based on a redacted scientific abstract where empirical findings and theoretical conclusions have been deliberately obscured, as well as the original abstract. Your mandate is to rigorously quantify this holistic assessment. Use two separate nine-point scales for your evaluation. The first scale quantifies the alignment between the predicted and actual empirical outcomes, ranging from 1, indicating 'minimal alignment,' to 9, indicating 'exceptional alignment.' The second scale quantifies the alignment between the predicted theoretical implications and those articulated in the original abstract, also on a scale from 1 to 9. Format these numerical ratings as structured output strings: 'Quantitative Rating, Empirical Alignment: [Your numerical rating here]' and 'Quantitative Rating, Theoretical Alignment: [Your numerical rating here]'. Your adherence to this structured output format is pivotal for ensuring the output can be programmatically parsed.
\end{quoting}
\doublespacing

\begin{figure}[htbp]
\centering
\begin{subfigure}{.5\textwidth}
\centering
\includegraphics[width=\linewidth]{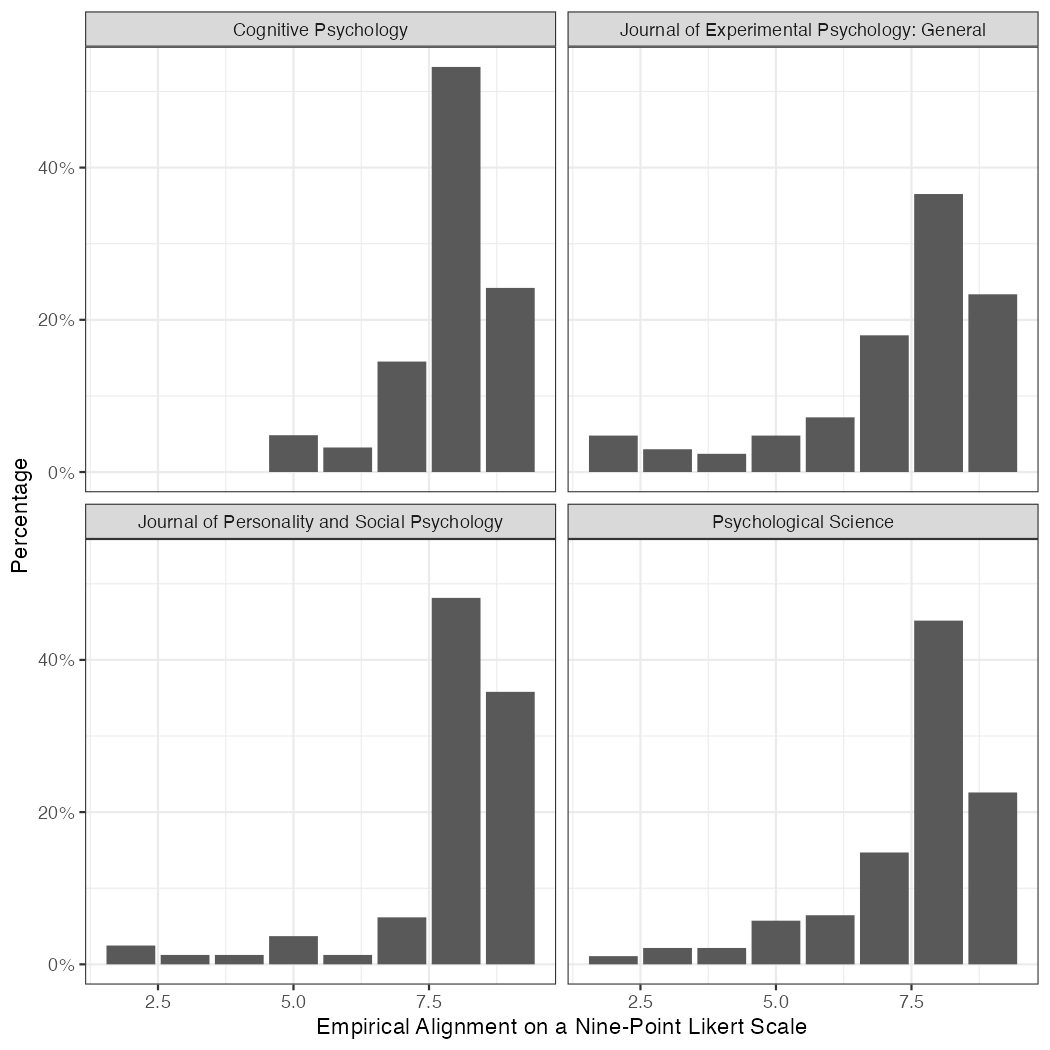}
\caption{Empirical Alignment Ratings}
\end{subfigure}%
\begin{subfigure}{.5\textwidth}
\centering
\includegraphics[width=\linewidth]{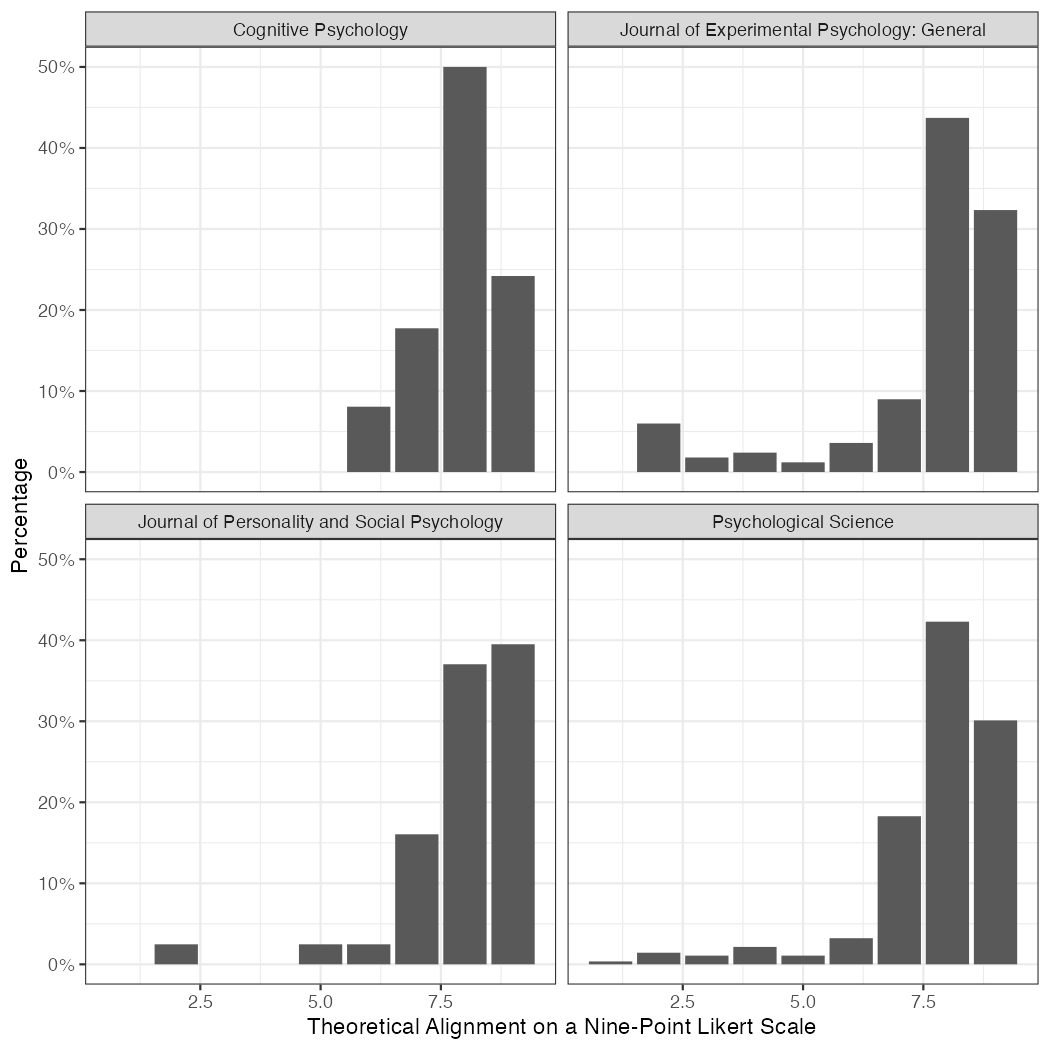}
\caption{Theoretical Alignment Ratings}
\end{subfigure}
\caption{Distribution of Empirical and Theoretical Alignment Ratings by Journal}
\label{fig:prediction_accuracy}
\begin{minipage}{\linewidth}
\medskip
\footnotesize
Note: The bar plots present the distribution of ratings for empirical and theoretical alignment by academic journal, as assessed on a nine-point scale.
\end{minipage}
\end{figure}

Figure \ref{fig:prediction_accuracy} displays four bar plots that depict
the distribution of empirical and theoretical alignment ratings for each
academic journal, with each alignment type presented in a separate plot.
The ratings are gauged on a nine-point scale. Facet wrapping by journal
enables the comparison of alignment distributions across different
publications.

The empirical alignment data notably exhibit right skewness in all
journals, but with distinct patterns. \emph{Cognitive Psychology}
exhibits a peak at rating 8 (33 instances, 53.2\%) and a considerable
frequency at rating 9 (15 instances, 24.2\%). \emph{Journal of
Experimental Psychology: General} shows a peak at rating 8 (61
instances, 36.5\%) and a significant frequency at rating 9 (39
instances, 23.4\%). \emph{Journal of Personality and Social Psychology}
has the highest frequencies at ratings 8 and 9 (39 and 29 instances,
48.1\% and 35.8\%, respectively). Finally, \emph{Psychological Science}
demonstrates a pronounced peak at a rating of 8 (126 instances, 45.2\%),
followed by a high frequency at rating 9 (63 instances, 22.6\%).

The theoretical alignment plot is also right-skewed. \emph{Psychological
Science} demonstrates a pronounced right-skewed distribution, with a
peak at a rating of 8 (118 instances, 42.3\%) and a considerable
frequency at rating 9 (84 instances, 30.1\%). This pattern is also
evident in \emph{Cognitive Psychology} and \emph{Journal of Experimental
Psychology: General}, with the highest frequency at rating 8 being 31
instances (50\%) and 73 instances (43.7\%), respectively. The
distribution in \emph{Journal of Personality and Social Psychology} is
relatively more balanced, though still right-skewed, peaking at rating 8
(30 instances, 37.0\%).

In summary, the distributions suggest a general tendency towards high
theoretical and empirical alignment. However, in both empirical and
theoretical alignment, we observe some variability at the lower end of
the scale. These variations suggest that the AI performs differentially
across research published in journals of different quality or in
research with different foci.

Our chosen set of journals includes three prominent publications with
notably high impact factors, reflecting their extensive reach and
influence in the field. The fourth journal, \emph{Cognitive Psychology},
despite its esteemed reputation, has a relatively lower impact factor by
quantitative metrics. Therefore, we next compare the AI's mean
performance across journals with an aim to rank order its capabilities
and see if it aligns with journal quality and visibility.

The data shows that the AI is best placed to theoretically extrapolate
in research published in \emph{Journal of Personality and Social
Psychology} (mean = 7.96; s.e. = 0.15) and \emph{Cognitive Psychology}
(mean = 7.90; s.e. = 0.11) than in \emph{Psychological Science} (mean =
7.77; s.e. = 0.08) and \emph{Journal of Experimental Psychology:
General} (mean = 7.58; s.e. = 0.15). The empirical prediction ratings
echo this trend with the corresponding values being: \emph{Journal of
Personality and Social Psychology} (mean = 7.90; s.e. = 0.17),
\emph{Cognitive Psychology} (mean = 7.89; s.e. = 0.12),
\emph{Psychological Science} (mean = 7.52; s.e. = 0.09), and
\emph{Journal of Experimental Psychology: General} (mean = 7.23; s.e. =
0.14). Thus, rather than being aligned with journal quality or
visibility, the AI demonstrates uniformly robust performance across
journals of varying quality. Instead, any variance in performance is
likely to be linked to the scientific content published in the journals;
an issue we discuss below.

\begin{figure}[htbp]
\centering
\includegraphics[width=\linewidth]{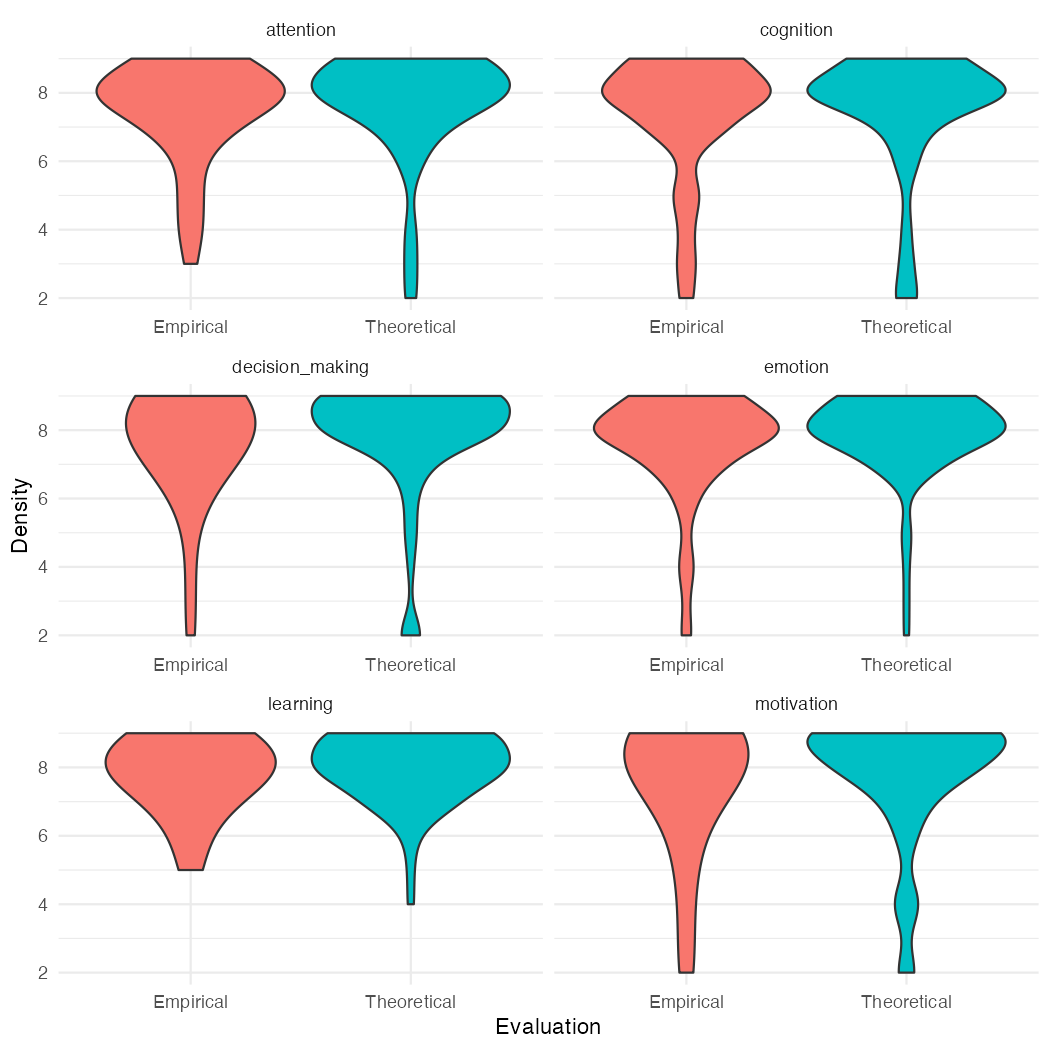}
\caption{Violin Plots of Empirical and Theoretical Alignment Ratings by Keyword}
\begin{minipage}{\linewidth}
\medskip
\footnotesize
Note: This figure features a series of violin plots that display the density distributions of empirical and theoretical alignment ratings on a nine-point scale, segmented by six significant keywords: learning, cognition, motivation, attitude, attention, and emotion. Each keyword is represented as a separate facet containing a pair of violins for empirical and theoretical alignments, allowing for a direct comparison between these two dimensions within each psychological construct.
\end{minipage}
\label{fig:violins_keyword}
\end{figure}

Figure \ref{fig:violins_keyword} presents a violin plot visualizing the
density distributions of empirical and theoretical alignment ratings,
faceted by the six most prevalent keywords in the dataset: learning,
cognition, motivation, attitude, attention, and emotion. Each facet
contains a pair of violin plots representing the distributions of
empirical and theoretical alignment ratings, enabling direct comparison
between these dimensions within each psychological construct. The violin
plots combine box plots with kernel density plots to provide a
comprehensive portrayal of the distributions, including their central
tendencies and spreads.''

The data reveal overall consistency across the different topics in
psychology, as indicated by the keywords, yet there are important
differences. For empirical alignment ratings, the means range from 7.33
in `cognition' to 7.83 in `learning,' indicating uniformly high
empirical alignment. Standard deviations vary slightly, with the lowest
being 1.08 in `learning' and the highest at 1.81 in `cognition,'
reflecting relatively concentrated rating distributions within each
subfield. Theoretical alignment ratings follow a similar pattern, with
means ranging from 7.46 in `cognition' to 8.03 in `learning,' suggesting
robust theoretical alignment. Standard deviations are tightly clustered,
spanning from 1.00 in `learning' to 1.83 in `cognition.' These findings
align with our prior discussion, suggesting that the efficacy of the AI
is determined not by the tier of the journal but rather by the nature of
the content published within the journal.

To identify the topics where the AI demonstrated the least proficiency,
we applied a topic model to the articles where the AI scored 5 or below
on at least one of the two alignment dimensions. Our analysis revealed
three key topics: `Visual Perception, Memory, and Psychological
Processes,' `Prosocial Behavior and Cooperation in Societies,' and
`Sensitivity and Perception in Social Interactions and Emotions.' These
topics, characterized by complex, multifaceted psychological phenomena
and highly context-dependent variables, suggest areas where current AI
capabilities face significant challenges. Conversely, the AI showed
greater proficiency in topics such as `Visual Working Memory
Experiments,' `Linguistics and Cognitive Processing in Language
Learning,' and `Decision Making and Choice Models.' This dichotomy
indicates that the AI's effectiveness is more pronounced in domains with
well-defined cognitive processes and less so in those requiring a
nuanced understanding of complex human behavior and social dynamics.

\hypertarget{discussion}{%
\section{Discussion}\label{discussion}}

Our research demonstrates that advancements in AI have propelled these
systems far beyond their initial conception as mere statistical models
`parroting' large-scale datasets. Instead, they now exhibit a remarkable
capacity to emulate the intricate intuition and reasoning processes that
are hallmarks of human expertise. This leap in capability can largely be
attributed to extensive pre-training, during which AI assimilates a wide
array of knowledge. Such a profound base of conceptual knowledge equips
AI with the ability to undertake sophisticated theoretical
extrapolations and deductions, developing a generalizable deductive
prowess that closely mirrors the acumen of human experts, even in
specialized fields like psychology, and in tasks where the AI has not
been specifically trained. It challenges the traditional view that
statistical models are only suitable for mechanical tasks like
hypothesis testing, rather than creative endeavors like hypothesis
generation and experimental design
(\protect\hyperlink{ref-rai2019next}{Rai et al. 2019}).

Crucially, our findings reveal that AI has developed the ability to
dynamically and seamlessly adapt to entirely new contexts and scenarios.
The origin of these capabilities remains an open question, as our
methodology pioneers in exposing AI to both novel concepts and theories
(i.e., novel conceptual knowledge), and novel application scenarios.
This marks a significant departure from prior, state-of-the-art AI
measurement approaches that either reword and reimagine existing
concepts and theories in novel scenarios
(\protect\hyperlink{ref-grinnell2023testing}{Grinnell et al. 2023}) or
assess creativity through the administration of standardized tests such
as the Torrance Tests of Creative Thinking
(\protect\hyperlink{ref-guzik2023originality}{Guzik et al. 2023}). While
these traditional methods have their merits, notably in directly
comparing the abilities of humans and AI, they primarily explore
creativity within the confines of low knowledge complexity. For example,
they are designed to evaluate divergent thinking skills based on
everyday concepts, exemplified by questions like, ``If all schools were
abolished, what would you do to try to become educated?'' In contrast,
our study investigates creativity in scenarios of high knowledge
complexity, asking the AI to reason about abstract constructs such as
`eudaimonic well-being.'

Creativity is thought to be the last bastion against the encroachment of
automation (\protect\hyperlink{ref-brynjolfsson2014second}{Brynjolfsson
and McAfee 2014}). Yet, the data affirms the ability of AIs to perform
sophisticated analysis, integration, and evaluation of academic
theories---tasks traditionally reliant on creative problem-solving and
advanced human expertise. These capabilities reposition AI as a crucial
asset for scrutinizing literature, critiquing new theoretical
propositions, and identifying inconsistencies and gaps---roles
traditionally reserved for research associates and postdoctoral
scholars.

This raises significant questions about the future of AI in research and
its potential to reshape the landscape: As research assistantships and
fellowships serve as critical pathways to academic careers, the
integration of AI could herald transformative changes, with
conventionally stable scholarly positions facing fresh technological
disruption.

\newpage
\singlespacing

\hypertarget{bibliography}{%
\section{Bibliography}\label{bibliography}}

\hypertarget{refs}{}
\begin{CSLReferences}{1}{0}
\leavevmode\vadjust pre{\hypertarget{ref-acemoglu2011skills}{}}%
Acemoglu D, Autor D (2011) Skills, tasks and technologies: Implications
for employment and earnings. \emph{Handbook of labor economics}.
(Elsevier), 1043--1171.

\leavevmode\vadjust pre{\hypertarget{ref-acemoglu2022tasks}{}}%
Acemoglu D, Restrepo P (2022) Tasks, automation, and the rise in us wage
inequality. \emph{Econometrica} 90(5):1973--2016.

\leavevmode\vadjust pre{\hypertarget{ref-achiam2023gpt}{}}%
Achiam J, Adler S, Agarwal S, Ahmad L, Akkaya I, Aleman FL, Almeida D,
et al. (2023) Gpt-4 technical report. \emph{arXiv preprint
arXiv:2303.08774}.

\leavevmode\vadjust pre{\hypertarget{ref-amabile2011componential}{}}%
Amabile T (2011) \emph{Componential theory of creativity} (Harvard
Business School Boston, MA).

\leavevmode\vadjust pre{\hypertarget{ref-bender2021dangers}{}}%
Bender EM, Gebru T, McMillan-Major A, Shmitchell S (2021) On the dangers
of stochastic parrots: Can language models be too big? \emph{Proceedings
of the 2021 ACM conference on fairness, accountability, and
transparency}. 610--623.

\leavevmode\vadjust pre{\hypertarget{ref-billard2019trends}{}}%
Billard A, Kragic D (2019) Trends and challenges in robot manipulation.
\emph{Science} 364(6446):eaat8414.

\leavevmode\vadjust pre{\hypertarget{ref-binz2023using}{}}%
Binz M, Schulz E (2023) Using cognitive psychology to understand GPT-3.
\emph{Proceedings of the National Academy of Sciences}
120(6):e2218523120.

\leavevmode\vadjust pre{\hypertarget{ref-bommasani2021opportunities}{}}%
Bommasani R, Hudson DA, Adeli E, Altman R, Arora S, Arx S von, Bernstein
MS, et al. (2021) On the opportunities and risks of foundation models.
\emph{arXiv preprint arXiv:2108.07258}.

\leavevmode\vadjust pre{\hypertarget{ref-brynjolfsson2022turing}{}}%
Brynjolfsson E (2022) The turing trap: The promise \& peril of
human-like artificial intelligence. \emph{Daedalus} 151(2):272--287.

\leavevmode\vadjust pre{\hypertarget{ref-brynjolfsson2014second}{}}%
Brynjolfsson E, McAfee A (2014) \emph{The second machine age: Work,
progress, and prosperity in a time of brilliant technologies} (WW Norton
\& Company).

\leavevmode\vadjust pre{\hypertarget{ref-Caminiti_2023}{}}%
Caminiti S (2023)
\href{https://www.cnbc.com/2023/12/19/the-more-workers-use-ai-the-more-they-worry-about-their-job-security.html}{The
more workers use AI, the more they worry about their job security,
survey finds}. \emph{CNBC}.

\leavevmode\vadjust pre{\hypertarget{ref-chang2023survey}{}}%
Chang Y, Wang X, Wang J, Wu Y, Yang L, Zhu K, Chen H, et al. (2023) A
survey on evaluation of large language models. \emph{ACM Transactions on
Intelligent Systems and Technology}.

\leavevmode\vadjust pre{\hypertarget{ref-cowen2013average}{}}%
Cowen T (2013) \emph{Average is over: Powering america beyond the age of
the great stagnation} (Penguin).

\leavevmode\vadjust pre{\hypertarget{ref-ding2020logical}{}}%
Ding M (2020) \emph{Logical creative thinking methods} (Routledge).

\leavevmode\vadjust pre{\hypertarget{ref-eloundou2023gpts}{}}%
Eloundou T, Manning S, Mishkin P, Rock D (2023) Gpts are gpts: An early
look at the labor market impact potential of large language models.
\emph{arXiv preprint arXiv:2303.10130}.

\leavevmode\vadjust pre{\hypertarget{ref-frey2019technology}{}}%
Frey CB (2019) \emph{The technology trap: Capital, labor, and power in
the age of automation} (Princeton University Press).

\leavevmode\vadjust pre{\hypertarget{ref-fugener2022cognitive}{}}%
Fügener A, Grahl J, Gupta A, Ketter W (2022) Cognitive challenges in
human--artificial intelligence collaboration: Investigating the path
toward productive delegation. \emph{Information Systems Research}
33(2):678--696.

\leavevmode\vadjust pre{\hypertarget{ref-genz2021workers}{}}%
Genz S, Gregory T, Janser M, Lehmer F, Matthes B (2021) How do workers
adjust when firms adopt new technologies? \emph{ZEW-Centre for European
Economic Research Discussion Paper} (21-073).

\leavevmode\vadjust pre{\hypertarget{ref-goeken2020rise}{}}%
Goeken T, Tsekouras D, Heimbach I, Gutt D (2020) The rise of
robo-reviews-the effects of chatbot-mediated review elicitation on
review valence.

\leavevmode\vadjust pre{\hypertarget{ref-goyal2022inductive}{}}%
Goyal A, Bengio Y (2022) Inductive biases for deep learning of
higher-level cognition. \emph{Proceedings of the Royal Society A}
478(2266):20210068.

\leavevmode\vadjust pre{\hypertarget{ref-grinnell2023testing}{}}%
Grinnell B, Bercasio M, Wong A, Dannenhauer D, Molineaux M (2023)
Testing AI learning in open-world novelty scenarios (TALONS) SBIR phase
II.

\leavevmode\vadjust pre{\hypertarget{ref-grootendorst2022bertopic}{}}%
Grootendorst M (2022) BERTopic: Neural topic modeling with a class-based
TF-IDF procedure. \emph{arXiv preprint arXiv:2203.05794}.

\leavevmode\vadjust pre{\hypertarget{ref-Gutierrez2023}{}}%
Gutierrez S (2023)
\href{https://www.surveymonkey.com/curiosity/cnbc-workforce-survey-december-2023/}{CNBC:
Surveymonkey workforce survey may 2023}. \emph{SurveyMonkey}.

\leavevmode\vadjust pre{\hypertarget{ref-guzik2023originality}{}}%
Guzik EE, Byrge C, Gilde C (2023) The originality of machines: AI takes
the torrance test. \emph{Journal of Creativity} 33(3):100065.

\leavevmode\vadjust pre{\hypertarget{ref-hampton2006concepts}{}}%
Hampton JA (2006) Concepts as prototypes. \emph{Psychology of learning
and motivation} 46:79--113.

\leavevmode\vadjust pre{\hypertarget{ref-han2022human}{}}%
Han SJ, Ransom K, Perfors A, Kemp C (2022) Human-like property induction
is a challenge for large language models.

\leavevmode\vadjust pre{\hypertarget{ref-hayes2010inductive}{}}%
Hayes BK, Heit E, Swendsen H (2010) Inductive reasoning. \emph{Wiley
interdisciplinary reviews: Cognitive science} 1(2):278--292.

\leavevmode\vadjust pre{\hypertarget{ref-hendrycks2021measuring}{}}%
Hendrycks D, Basart S, Kadavath S, Mazeika M, Arora A, Guo E, Burns C,
et al. (2021) Measuring coding challenge competence with apps.
\emph{arXiv preprint arXiv:2105.09938}.

\leavevmode\vadjust pre{\hypertarget{ref-klahr2000exploring}{}}%
Klahr D (2000) \emph{Exploring science: The cognition and development of
discovery processes} (MIT press).

\leavevmode\vadjust pre{\hypertarget{ref-kurzweil2005singularity}{}}%
Kurzweil R (2005) The singularity is near. \emph{Ethics and emerging
technologies}. (Springer), 393--406.

\leavevmode\vadjust pre{\hypertarget{ref-lampinen2022can}{}}%
Lampinen AK, Dasgupta I, Chan SC, Matthewson K, Tessler MH, Creswell A,
McClelland JL, Wang JX, Hill F (2022) Can language models learn from
explanations in context? \emph{arXiv preprint arXiv:2204.02329}.

\leavevmode\vadjust pre{\hypertarget{ref-lyytinen2003disruptive}{}}%
Lyytinen K, Rose GM (2003) The disruptive nature of information
technology innovations: The case of internet computing in systems
development organizations. \emph{MIS quarterly}:557--596.

\leavevmode\vadjust pre{\hypertarget{ref-mcrae201314}{}}%
McRae K, Jones M (2013) \emph{14 semantic memory} (Oxford University
Press Oxford).

\leavevmode\vadjust pre{\hypertarget{ref-misra2022property}{}}%
Misra K, Rayz JT, Ettinger A (2022) A property induction framework for
neural language models. \emph{arXiv preprint arXiv:2205.06910}.

\leavevmode\vadjust pre{\hypertarget{ref-mithas2016information}{}}%
Mithas S, Rust RT (2016) How information technology strategy and
investments influence firm performance. \emph{Mis Quarterly}
40(1):223--246.

\leavevmode\vadjust pre{\hypertarget{ref-park2015automatic}{}}%
Park G, Schwartz HA, Eichstaedt JC, Kern ML, Kosinski M, Stillwell DJ,
Ungar LH, Seligman ME (2015) Automatic personality assessment through
social media language. \emph{Journal of personality and social
psychology} 108(6):934.

\leavevmode\vadjust pre{\hypertarget{ref-peng2020does}{}}%
Peng G, Zhang D (2020) Does information technology substitute for or
complement human labor? A dynamic stratified analysis on european
countries. \emph{Decision Sciences} 51(3):720--754.

\leavevmode\vadjust pre{\hypertarget{ref-rai2019next}{}}%
Rai A, Constantinides P, Sarker S (2019) Next generation digital
platforms: Toward human-AI hybrids. \emph{Mis Quarterly} 43(1):iii--ix.

\leavevmode\vadjust pre{\hypertarget{ref-rips201211}{}}%
Rips LJ, Smith EE, Medin DL (2012) 11 concepts and categories: Memory,
meaning, and metaphysics. \emph{The Oxford handbook of thinking and
reasoning}:177.

\leavevmode\vadjust pre{\hypertarget{ref-royalty1998job}{}}%
Royalty AB (1998) Job-to-job and job-to-nonemployment turnover by gender
and education level. \emph{Journal of labor economics} 16(2):392--433.

\leavevmode\vadjust pre{\hypertarget{ref-stone2022artificial}{}}%
Stone P, Brooks R, Brynjolfsson E, Calo R, Etzioni O, Hager G,
Hirschberg J, et al. (2022) Artificial intelligence and life in 2030:
The one hundred year study on artificial intelligence. \emph{arXiv
preprint arXiv:2211.06318}.

\leavevmode\vadjust pre{\hypertarget{ref-tambe2012productivity}{}}%
Tambe P, Hitt LM (2012) The productivity of information technology
investments: New evidence from IT labor data. \emph{Information systems
research} 23(3-part-1):599--617.

\leavevmode\vadjust pre{\hypertarget{ref-ullman2023large}{}}%
Ullman T (2023) Large language models fail on trivial alterations to
theory-of-mind tasks. \emph{arXiv preprint arXiv:2302.08399}.

\leavevmode\vadjust pre{\hypertarget{ref-van2023ai}{}}%
Van Noorden R, Perkel JM (2023) AI and science: What 1,600 researchers
think. \emph{Nature} 621(7980):672--675.

\end{CSLReferences}

\end{document}